\documentclass[nohyperref]{article}

\usepackage{microtype}
\usepackage{graphicx}
\usepackage{subcaption}
\usepackage{booktabs} 
\usepackage{multirow}
\usepackage{color, colortbl}
\usepackage{hyperref}
\usepackage{bm}
\usepackage{bbm}
\usepackage{pifont}
\usepackage[linesnumbered,ruled,vlined]{algorithm2e}
\definecolor{codeblue}{rgb}{0.25,0.5,0.5}

\SetCommentSty{mycommfont}
\definecolor{Gray}{gray}{0.9}
\def\*#1{\mathbf{#1}}

\newcommand{\cmark}{\ding{51}}%
\newcommand{\xmark}{\ding{55}}%

\usepackage[accepted]{icml2022}

\usepackage{amsmath}
\usepackage{amssymb}
\usepackage{mathtools}
\usepackage{amsthm}
\usepackage{graphicx}
\usepackage{subcaption}
\usepackage{caption}
\usepackage{wrapfig}
\usepackage[capitalize,noabbrev]{cleveref}

\theoremstyle{plain}
\newtheorem{theorem}{Theorem}[section]

\newtheorem{lemma}[theorem]{Lemma}

\theoremstyle{definition}

\theoremstyle{remark}

\usepackage[textsize=tiny]{todonotes}

\icmltitlerunning{POEM: Out-of-Distribution Detection with Posterior Sampling}

\begin{document}

\twocolumn[
\icmltitle{POEM: Out-of-Distribution Detection with Posterior Sampling}



\icmlsetsymbol{equal}{*}

\begin{icmlauthorlist}
\icmlauthor{Yifei Ming}{equal,yyy}
\icmlauthor{Ying  Fan}{equal,yyy}
\icmlauthor{Yixuan Li}{yyy}
\end{icmlauthorlist}

\icmlaffiliation{yyy}{Department of Computer Sciences, University of Wisconsin-Madison, USA}

\icmlcorrespondingauthor{Yifei Ming}{ming5@wisc.edu}
\icmlcorrespondingauthor{Ying Fan}{yfan87@wisc.edu}
\icmlcorrespondingauthor{Yixuan Li}{sharonli@cs.wisc.edu}

\icmlkeywords{Machine Learning, ICML}

\vskip 0.3in
]



\printAffiliationsAndNotice{\icmlEqualContribution} 

\begin{abstract}
  Out-of-distribution (OOD) detection is indispensable for machine learning models deployed in the open world. 
  Recently, the use of an auxiliary outlier dataset during training (also known as outlier exposure) has shown promising performance. As the sample space for potential OOD data can be prohibitively large, sampling informative outliers is essential.
  In this work, we propose a novel posterior sampling-based outlier mining framework, POEM, which facilitates efficient use of outlier data and promotes learning a compact decision boundary between ID and OOD data for improved detection. We show that POEM establishes \emph{state-of-the-art} performance on common benchmarks. Compared to the current best method that uses a greedy sampling strategy, POEM improves the relative performance by {42.0}\% and {24.2}\% (FPR95) on CIFAR-10 and CIFAR-100, respectively. We further provide theoretical insights on the effectiveness of POEM for OOD detection. 
\end{abstract}

\begin{figure*}[!t]
 \begin{center}
      \begin{subfigure}[t]{0.18\linewidth}
    \includegraphics[width=\linewidth]{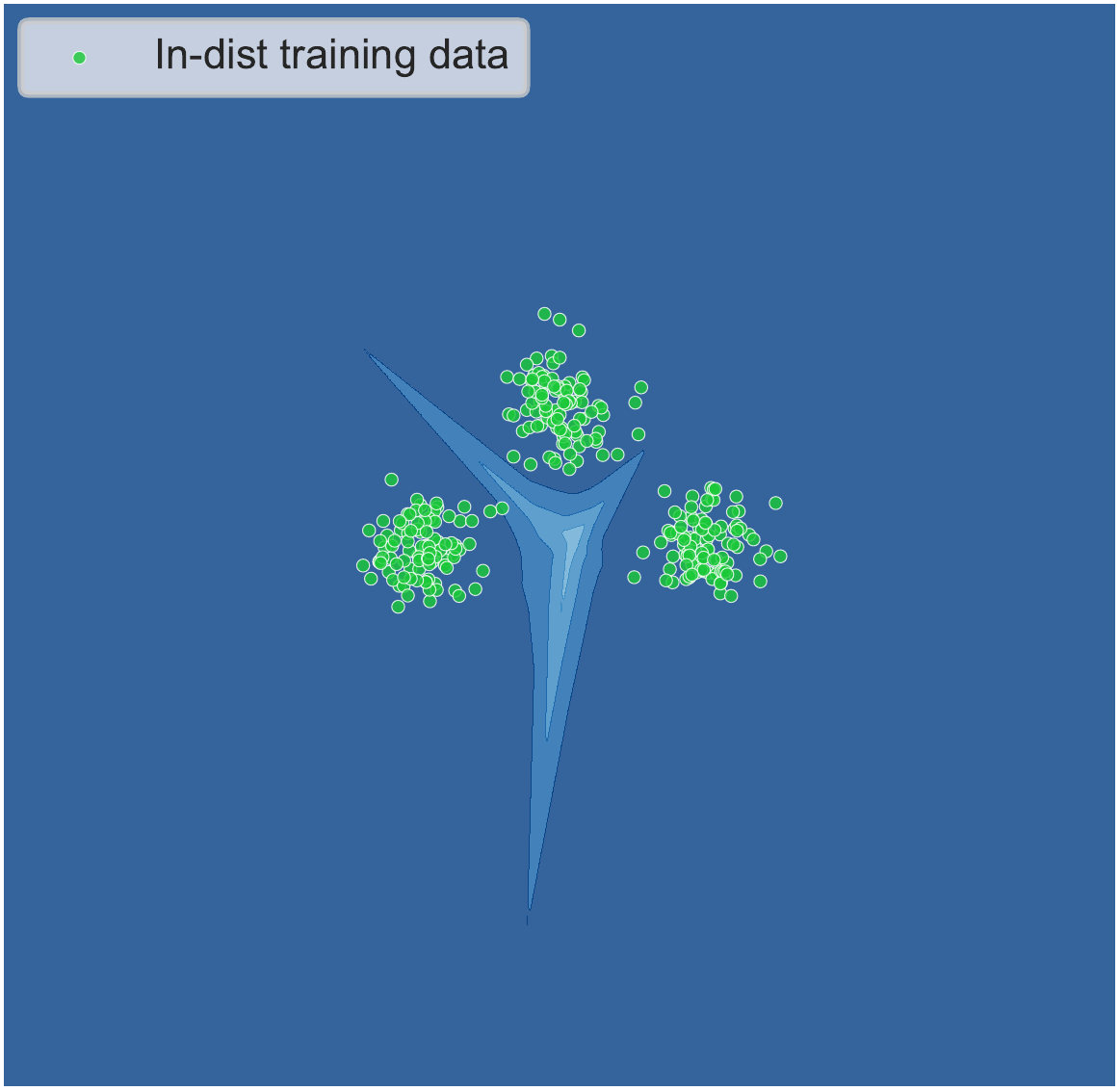}
     \caption{ID data only}
     \label{fig:overfit}
  \end{subfigure}
   \begin{subfigure}[t]{0.0318\linewidth}
    \includegraphics[width=\linewidth]{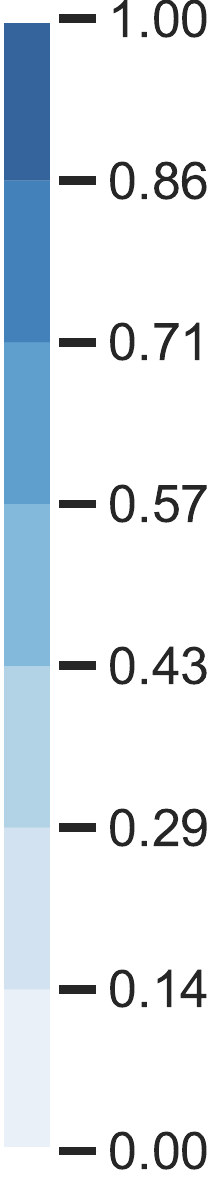}
  \end{subfigure}
  \begin{subfigure}[t]{0.19\linewidth}
    \includegraphics[width=\linewidth]{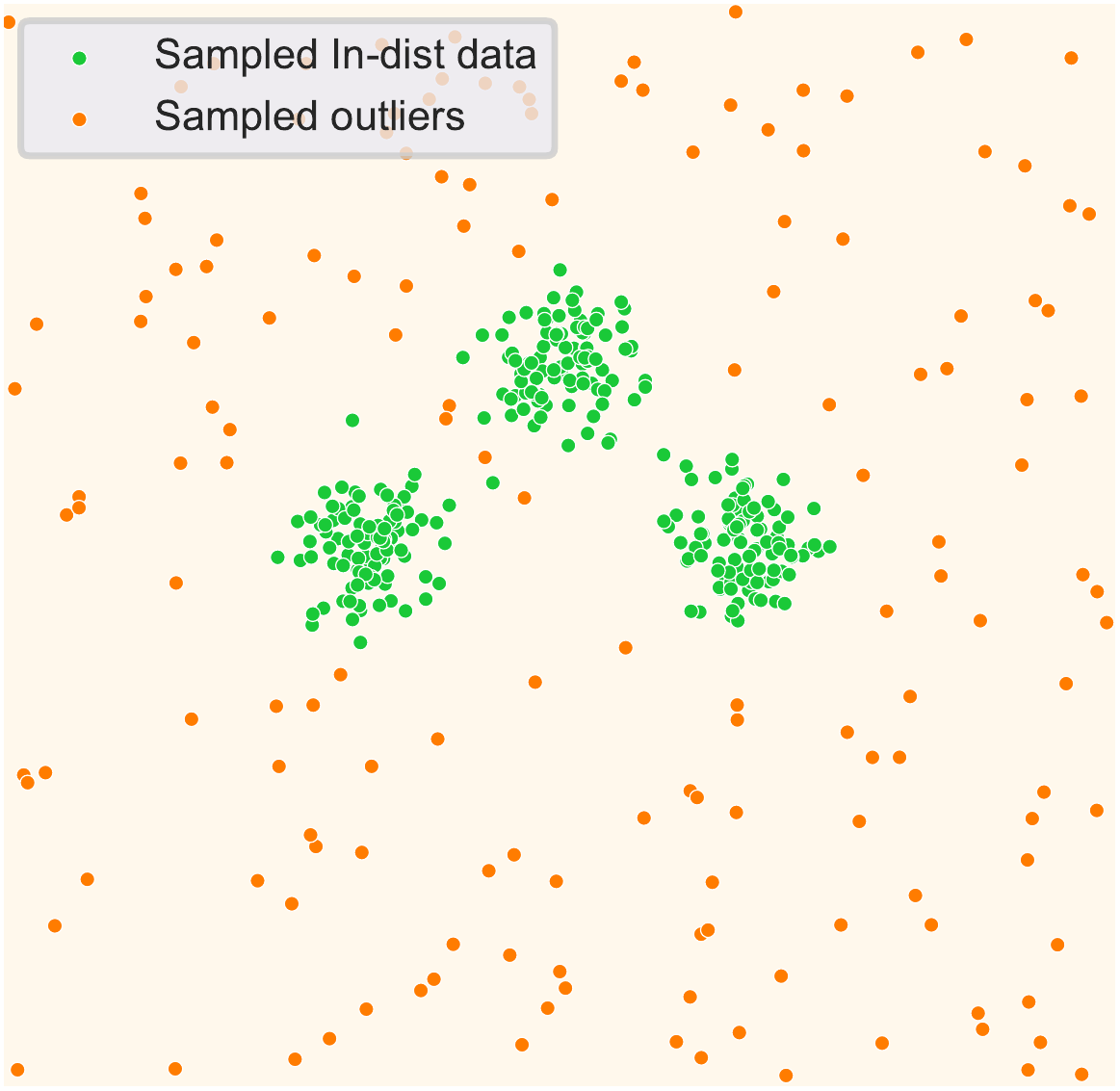}
     \caption{Sample Space}
     \label{fig:b}
  \end{subfigure}
  \begin{subfigure}[t]{0.19\linewidth}
    \includegraphics[width=\linewidth]{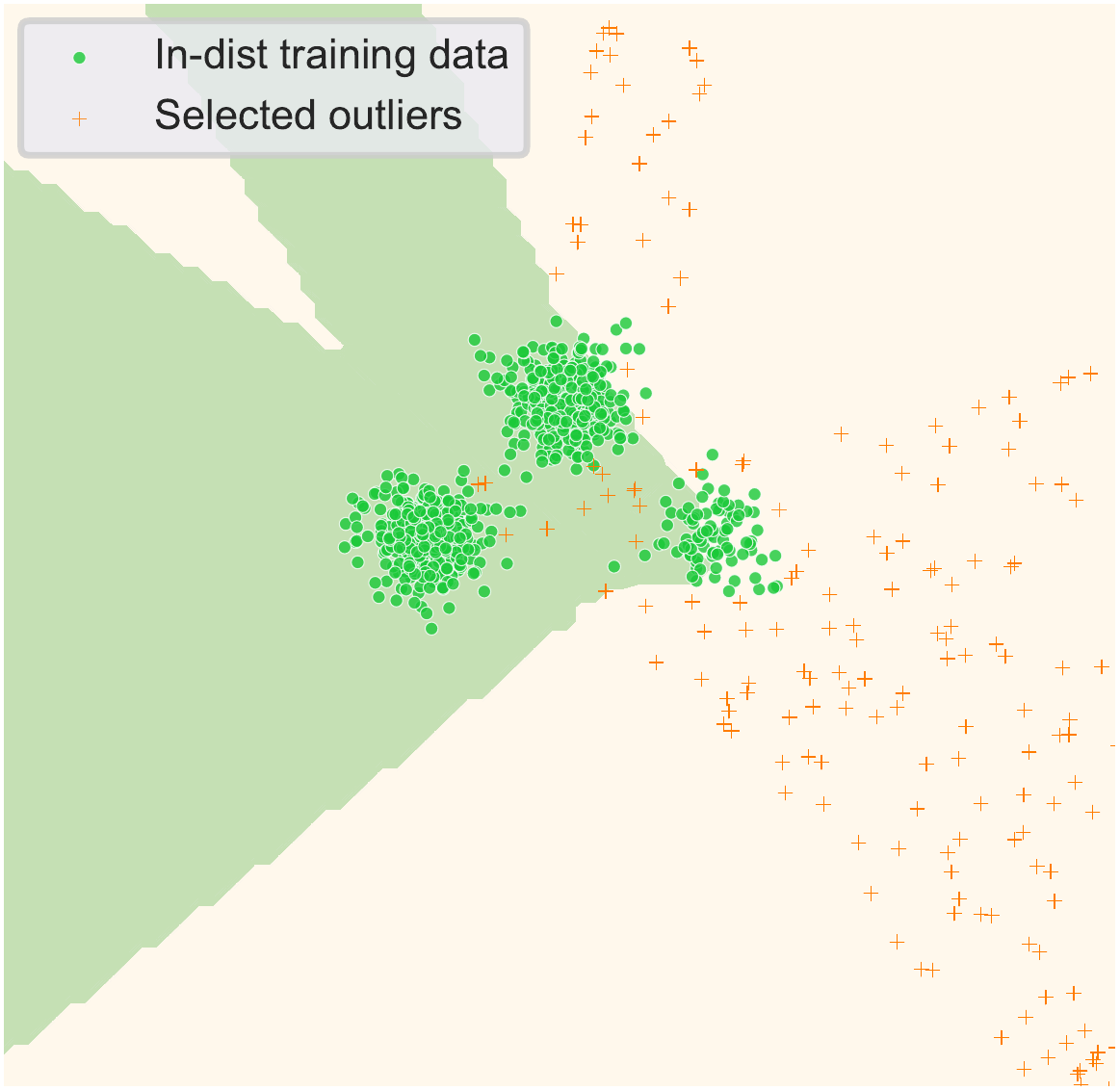}
    \caption{Epoch 1}
  \end{subfigure}
    \begin{subfigure}[t]{0.19\linewidth}
    \includegraphics[width=\linewidth]{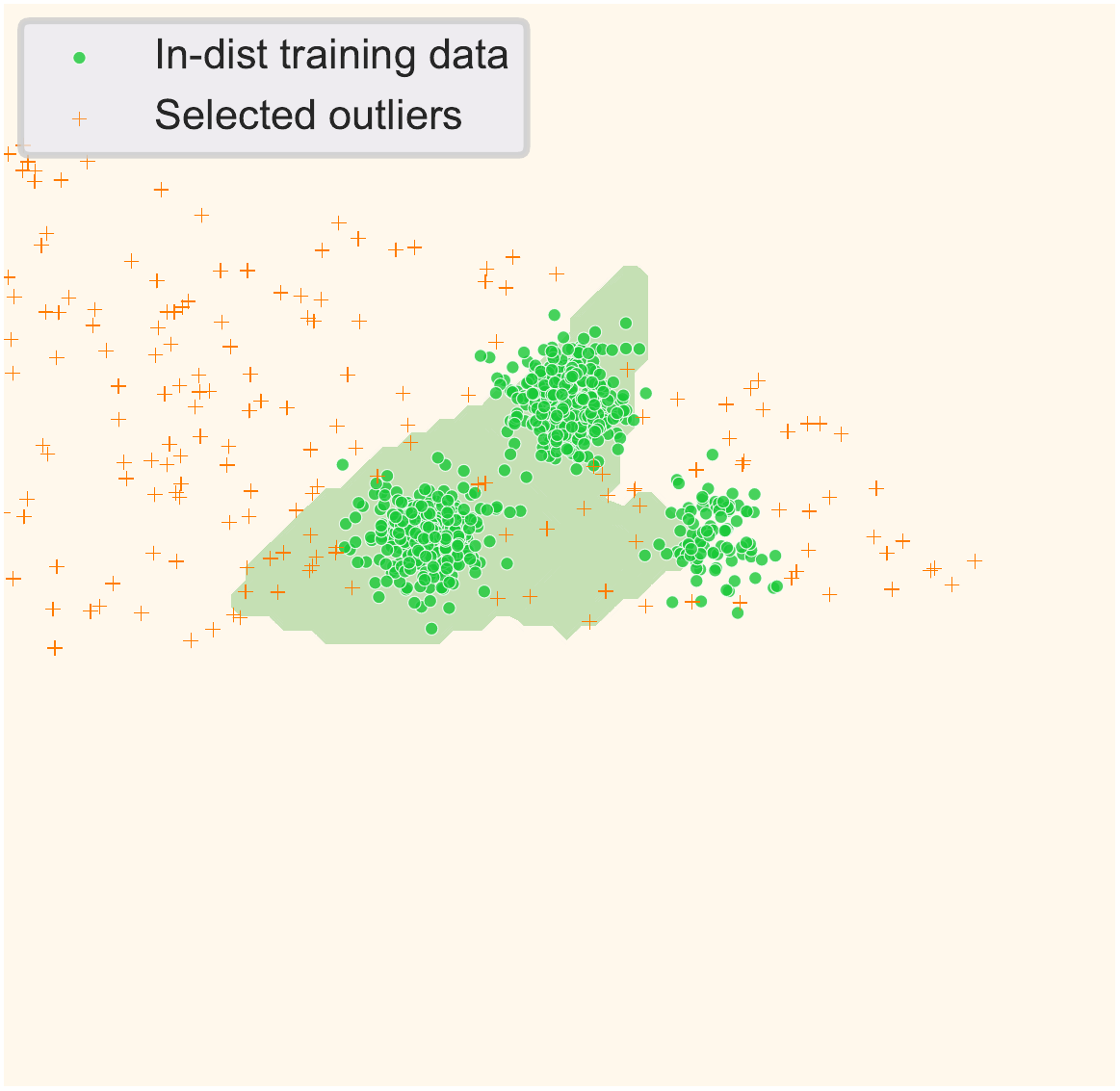}
    \caption{Epoch 4}
  \end{subfigure}
    \begin{subfigure}[t]{0.19\linewidth}
    \includegraphics[width=\linewidth]{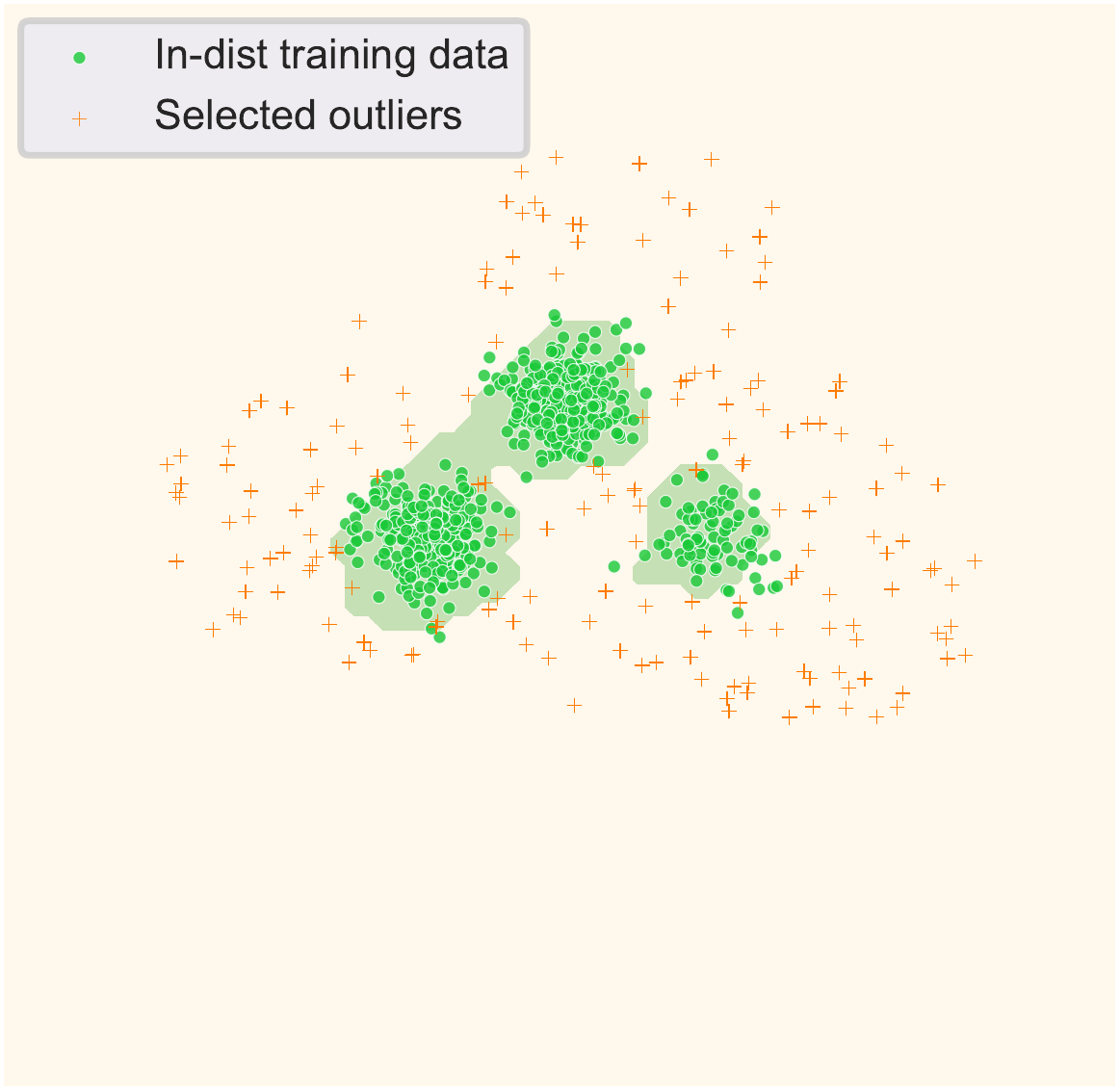}
    \caption{Epoch 30}
    \label{fig:d}
  \end{subfigure}
\caption{\small Illustration of our \emph{Posterior Sampling-based Outlier Mining}~\textbf{(POEM)} framework. (a) Samples from  in-distribution $\mathcal{P}_\text{in}$ (green dots). Blue shades represent the maximum predicted probability of a neural network classifier trained \emph{without} an auxiliary outlier dataset. The resulting classifier is overly confident for regions (dark blue) far away from the ID data. (b) Some samples drawn from $\mathcal{P}_\text{in}$ (green dots) and auxiliary outlier data $\mathcal{P}_\text{aux}$ (orange dots).  (c) to (e): Selected outliers (orange dots) from posterior distributions at different training epochs, along with the decision boundary (green for ID and beige for OOD) of the classifier.}
\vspace{-0.2cm}
\label{fig:teaser}
\end{center}
\end{figure*}

\section{Introduction}
Out-of-distribution (OOD) detection has become a central challenge in safely deploying machine learning models in the open world, where the test data can naturally arise from a different distribution. Concerningly, modern neural networks are shown to produce overconfident and therefore untrustworthy predictions for OOD inputs~\cite{fool}. To mitigate the issue, recent works have explored training with a large auxiliary outlier dataset, where the model is regularized to produce lower confidence~\cite{lee2018gan, OE, ccu, chen2020informative} or higher energy~\cite{liu2020energybased} on the outlier training data. These methods have demonstrated promising OOD detection performance over the counterpart (without auxiliary data). 

Despite encouraging results, existing methods suffer from ineffective use of outliers, and have largely underlooked the importance of {outlier mining}. In particular, the sample space for potential outlier data can be prohibitively large, making the majority of outliers uninformative for model regularization. As recently observed by Chen et~al.~\yrcite{chen2020informative}, \emph{randomly} selecting outlier samples during training yields a large portion of uninformative samples that do not meaningfully improve the estimated decision boundary between in-distribution (ID) and OOD data. The above limitations motivate the following important yet underexplored question: \emph{how can we efficiently utilize the auxiliary outlier data for model regularization}?

In this work, we propose a novel {Posterior Sampling-based Outlier Mining} \textbf{(POEM)} framework for OOD detection --- selecting the most informative outlier data from a large pool of auxiliary data points, which can help the model estimate a compact decision boundary between ID and OOD for improved OOD detection. 
For example, Figure~\ref{fig:teaser} shows the evolution of decision boundaries, as a result of outlier mining. Our novel idea is to formalize outlier mining as sequential decision making, where actions correspond to outlier selection, and the reward function is based on the closeness to the boundary between ID and OOD data.
In particular, we devise a novel reward function termed \emph{boundary score}, which is higher for outliers close to the boundary. 
Key to our framework, we leverage Thompson sampling~\cite{thompson1933likelihood}, also known as posterior sampling, to balance exploration and exploitation during reward optimization.

As an integral part of our framework, \texttt{POEM} maintains and updates the posterior over models to encourage better exploration. For efficient outlier mining, \texttt{POEM} chooses samples with the highest boundary scores based on the posterior distribution. Because the exact posterior is often intractable, we approximate it by performing Bayesian linear regression on top of feature representations extracted via a neural network~\cite{riquelme2018deep}. Posterior updates are performed periodically during training, which help shape the estimated decision boundary between ID and OOD accordingly (see Figure~\ref{fig:teaser}). Unlike the greedy mining approach~\cite{chen2020informative} that focuses on exploitation, \texttt{POEM} selects outliers via posterior sampling and allows for better balancing between exploration and exploitation. We show that our framework is computationally tractable and can be trained efficiently end-to-end in the context of modern deep neural networks.

We show that \texttt{POEM} demonstrates state-of-the-art OOD detection performance and enjoys good theoretical properties. On common OOD detection benchmarks, \texttt{POEM} significantly outperforms competitive baselines using randomly sampled outliers~\cite{OE, liu2020energybased, Self, ccu}. Compared to the greedy sampling strategy~\cite{chen2020informative}, \texttt{POEM} improves the relative performance by {42.0}\% and {24.2}\% (FPR95) on CIFAR-10 and CIFAR-100~\cite{krizhevsky2009learning} respectively, evaluated across six diverse OOD test datasets. Lastly, we provide theoretical analysis for \texttt{POEM}, and explain why outliers with high boundary scores benefit sample complexity.
Our main contributions are:
\begin{itemize}
\vspace{-0.15cm}
\item We propose a novel Posterior Sampling-based Outlier Mining framework (dubbed \texttt{POEM}), which facilitates efficient use of outlier data and promotes learning a compact decision boundary between ID and OOD data for improved OOD detection.
\vspace{-0.15cm}
\item We perform extensive experiments comparing \texttt{POEM} with competitive OOD detection methods and various sampling strategies. We show that \texttt{POEM} establishes \textbf{state-of-the-art} results on common benchmarks. \texttt{POEM} also displays strong performance with comparable computation as baselines via a tractable algorithm for maintaining and updating the posterior.
\vspace{-0.15cm}
\item   We provide theoretical insights on why outlier mining with high boundary scores benefits sample efficiency.
\end{itemize}

\vspace{-0.5cm}
\section{Preliminaries}
\label{class}
We consider supervised multi-class classification, where $\mathcal{X}=\mathbb{R}^d$ denotes the input space and $\mathcal{Y}=\{1,2,...,K\}$ denotes the label space. The training set $\mathcal{D}_{\text{in}}^{\text{train}} = \{(\*x_i, y_i)\}_{i=1}^N$ is drawn \emph{i.i.d.} from $P_{\mathcal{X}\mathcal{Y}}$. Let $\mathcal{P}_{\text{in}}$ denote the marginal (ID) distribution on $\mathcal{X}$.
 
 \vspace{-0.2cm}
\subsection{Out-of-distribution Detection} When deploying a model in the real world, a reliable classifier should not only accurately classify known in-distribution (ID) samples, but also identify as ``unknown'' any OOD input. This can be achieved through having dual objectives: OOD identification and multi-class classification of ID data~\cite{bendale2016towards}. 

OOD detection can be formulated as a binary classification problem.  At test time, the goal of OOD detection is to decide whether a sample $\*x \in \mathcal{X}$ is from $\mathcal{P}_\text{in}$ (ID) or not (OOD). In literature, OOD distribution $\mathcal{P}_\text{out}$ often simulates unknowns encountered during deployment time, such as samples from an irrelevant distribution {whose label set has no intersection with $\mathcal{Y}$ and therefore should not be predicted by the model}. The decision can be made via a thresholding comparison: 
\vspace{-0.3cm}
\begin{align*}
\label{eq:threshold}
	D_{\lambda}(\*x)=\begin{cases} 
      \text{ID} & S(\*x)\ge \gamma \\
      \text{OOD} & S(\*x) < \gamma 
  \end{cases},
\end{align*}
where samples with higher scores $S(\*x)$ are classified as ID and vice versa, and  $\gamma$ is the threshold. 

\textbf{Auxiliary outlier data.} While the test-time OOD distribution $\mathcal{P}_\text{out}$ remains unknown, recent works~\cite{OE} make use of an auxiliary unlabeled dataset $\mathcal{D}_\text{aux}$ drawn from $\mathcal{P}_\text{aux}$ for model regularization. The model is encouraged to be less confident in the auxiliary outliers. For terminology clarity, we refer to training-time examples as \emph{auxiliary outliers} and exclusively use \emph{OOD data} to refer to test-time unknown inputs.

Although previous works showed that utilizing an auxiliary outlier dataset during training improves OOD detection, 
they suffer from ineffective use of outliers due to the prohibitively large sample space of OOD data. This calls for a sample-efficient way of utilizing the auxiliary outlier data.

\section{Method}
\label{branch}

In this section, we present our novel framework, \emph{Posterior sampling-based outlier mining} (dubbed \textbf{POEM}) --- selecting the most informative outlier data from a large pool of unlabeled data points, which helps the model estimate a compact decision boundary between ID and OOD for improved OOD detection. Unlike using random sampling~\cite{OE, Self, liu2020energybased} or greedy sampling~\cite{chen2020informative}, our framework allows balancing exploitation and exploration, which enjoys both strong empirical performance and theoretical properties.

In designing POEM, we address three key challenges. We will first introduce how to model outlier mining via the Thompson sampling process (Section~\ref{sec:ts}), and then describe how to tractably maintain and update the posterior in the context of modern neural networks (Section~\ref{sec:posterior}). Lastly, we describe the training objective that incorporates the sampled outliers for model regularization (Section~\ref{sec:train}).

\begin{figure*}[htb]
  \begin{subfigure}[b]{0.7\linewidth}
    \includegraphics[width=\linewidth]{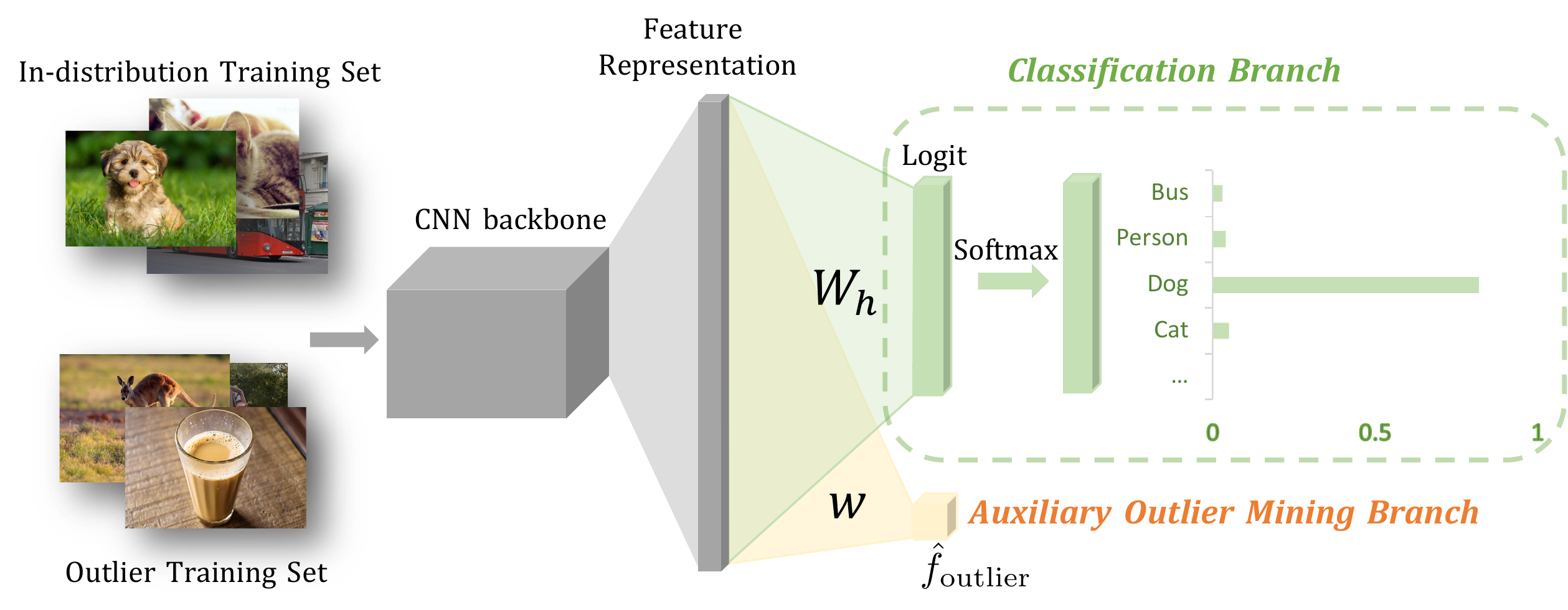}
     \caption{Method Overview}
     \label{fig_arch}
  \end{subfigure}
  \hfill
    \begin{subfigure}[b]{0.25\linewidth}
    \includegraphics[width=\linewidth]{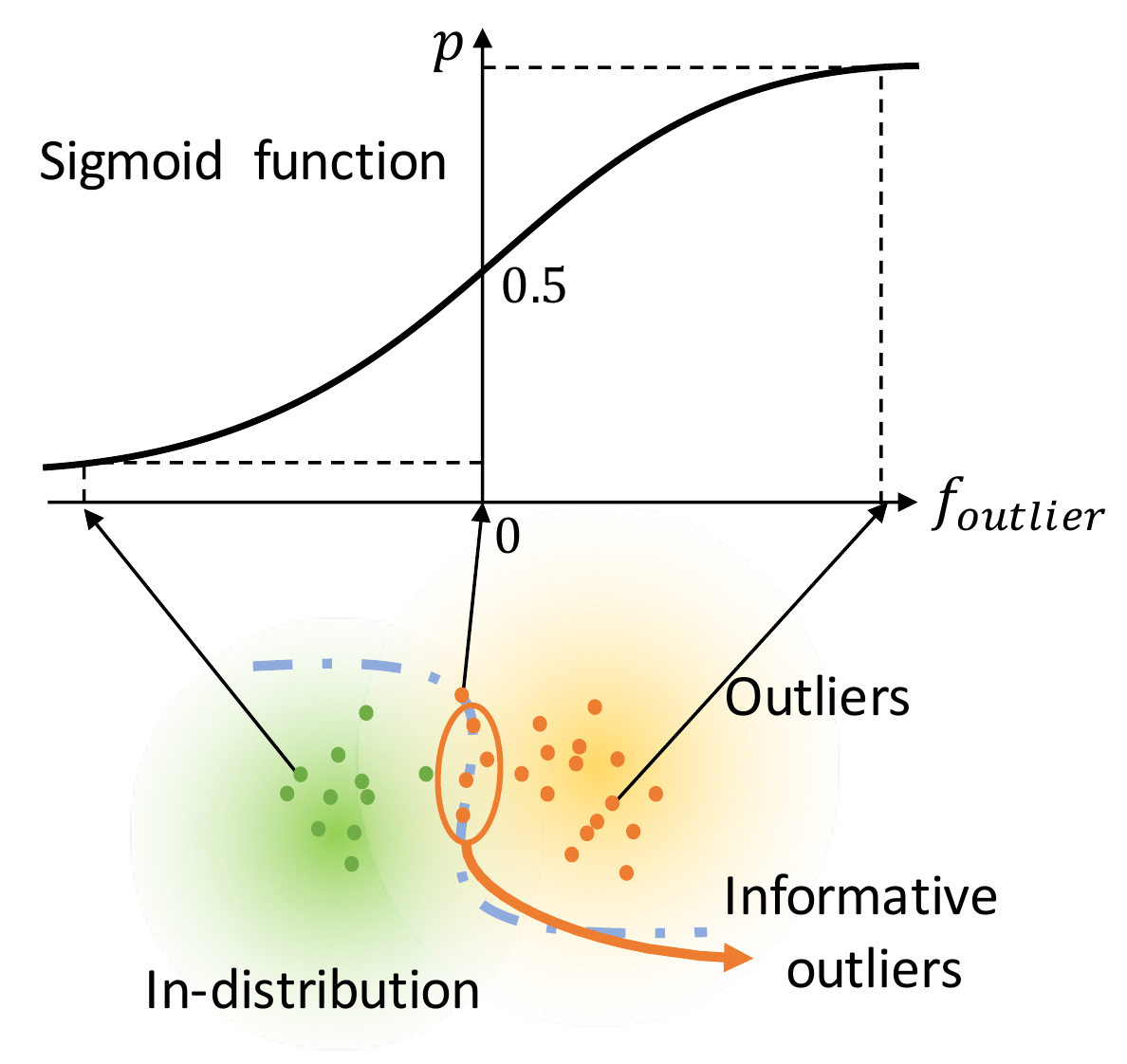}
    \caption{Boundary Score \& Density}
    \label{boundary}
  \end{subfigure}
  \vspace{-0.3cm}
\caption{ \small (a) Overview of \textbf{POEM}. We use a two-branch architecture with shared feature representation. During training, the outlier mining branch (beige) selects the near-boundary outliers. A mixture of outliers and ID data is used to train the classifier (green) and update the weights in the CNN backbone (grey). Based on the updated feature representation, the posterior update is applied to the weights $\*w$ of the Bayesian linear layer to refine the decision boundary between ID and outlier data. (b) The connection between the boundary score $G(\*x) = -|f_{\text{outlier}}(\mathbf{x};\*w^*)|$ and density. Outliers close to the boundary tend to have higher boundary scores.} 
\end{figure*}
\subsection{Outlier Mining: A Thompson Sampling View}
\label{sec:ts}

One novelty of this work is to formalize outlier mining as sequential decision making, where the actions correspond to outliers selection, and the reward function is based on the closeness to the (unknown) decision boundary between ID and OOD data. The goal of outlier mining is to identify the most informative outliers that are closer to the decision boundary between ID and OOD data. Key to our framework, we leverage the classic and elegant Thompson sampling algorithm~\cite{thompson1933likelihood} --- also known as posterior sampling --- to balance between  exploitation (\emph{i.e.}, maximizing immediate performance) and exploration (\emph{i.e.}, investing to accumulate
new information that may improve future performance). 

\textbf{A conceptual example of outlier mining.} To help readers understand the role of outlier mining, we provide a simple example in Figure~\ref{fig:teaser}. The in-distribution data (green) consists of three class-conditional Gaussians. Outlier training data (orange) is drawn from a uniform distribution and is at least two standard deviations away from the mean of every in-distribution class.
Figure~\ref{fig:teaser} (c)-(d) shows the evolution of decision boundaries between ID and outliers, as a result of outlier mining.
We can see that outlier mining significantly reduces the uninformative outliers selected from epoch 4 to epoch 30. These near-boundary outliers can help the network become aware of the real decision boundary between ID and outliers, which improves the OOD detection.

\paragraph{Formalize outlier mining.}
We now formalize the outlier mining process (see Algorithm~\ref{alg:TS} for an overview). At each step $t$, the model parameter $\*w_t$ is sampled from the posterior distribution, then the learner takes an action $a_t$ by choosing outlier $\mathbf{x} \sim \mathcal{P}_\text{aux}$ based on $\*w_t$.
The learner receives a reward
$G(\mathbf{x})$ termed \textbf{boundary score}, where a high boundary score indicates outliers being close to the boundary. The goal of outlier mining is to find the most informative outliers, \emph{i.e.}, those with the highest boundary scores, since they are more desirable for model regularization to learn a compact boundary between ID and OOD. A key component in Thompson sampling is to maintain a posterior over models to encourage better exploration. 
The posterior distribution is then updated after the reward is observed.

\begin{algorithm}[t]
  \caption{Outlier Mining via Thompson Sampling}
  \label{alg:TS}
\textbf{Input:} A prior distribution $P^{\*w}_0$ over $\*w$.

\begin{algorithmic}
 \FOR{step $t$ = $0,1,\cdots, T$ }
\STATE Sample $\*w_t\sim P^{\*w}_t$.
\STATE Take action $a_t$ by choosing outliers $\*x\sim \mathcal{P}_\text{aux}$ based on the sampled model $\*w_t$.
\STATE Receive some reward $G(\*x)$.
\STATE  Update the posterior distribution $P^{\*w}_{t+1}$ for model.
  \ENDFOR
\end{algorithmic}
\end{algorithm}

Concretely, we define $G(\*x) = -|f_{\text{outlier}}(\*x; \*w^*)|$, where $f_{\text{outlier}}$ is a function parameterized by some unknown ground truth weights $\*w^*$ and maps a high-dimensional input $\*x$ into a scalar. One can easily convert the logit to its probabilistic form by using the sigmoid function: $p(\text{outlier}|\*x) = \text{Sigmoid}(f_\text{outlier}(\*x; \*w^*))$. As shown in Figure \ref{boundary}, the near-boundary outliers would correspond to $|f_\text{outlier}(\*x; \*w^*)|\approx 0$. In a nutshell, we are optimizing an unknown function by selecting  samples, maintaining and modeling the distribution of $\*w$, and using this model to select near-boundary outliers over time. Unlike greedy algorithms that focus on exploitation, the Thompson sampling framework allows balancing the exploration and exploitation trade-off during the optimization. We proceed to describe details of maintaining and updating the posterior distribution for the model. 
 
{\LinesNumberedHidden
\begin{algorithm*}[t]
\small	
	\DontPrintSemicolon
	\SetNoFillComment
	\textbf{Input:} In-distribution training set $\mathcal{D}_{\text{in}}^{\text{train}} =\{\*x_i, y_i\}_{i=1}^N$ , a large outlier dataset $\mathcal{D}_{\text{aux}}$, neural network encoder $\phi$, weights $\*w_{\text{OM}}$ for Bayesian linear regression with prior covariance $\Sigma$, the variance of target logits noise $\sigma^2$, margin hyperparameters $m_{\text{in}}$ and $m_{\text{out}}$, regularizer weight $\beta$, pool size $S$, queue size $M$, prior distribution $\mathcal{N}(0, \Sigma)$ \\
	\\
	Sample $\*w_0 \sim  \mathcal{N}(0, \Sigma)$\\
	\For {$t=0,1,2,\ldots$}
	{
	Sample a subset of $\mathcal{D}_{\text{aux}}$ as the outlier pool $\mathcal{D}_{\text{pool}} = \{\*x_i\}_{i=1}^S$\\
	\For {$\*x_i \in \mathcal{D}_{\text{pool}}$}
	{
        Calculate the estimated boundary score $\hat{G}(\*x_i) = -|\hat{f}_\text{outlier}(\*x_i;\*w_t)| =-|\*w_t^\top \phi(\*x_i)|$
	}
	\tcp{select informative outliers}
	Construct $\mathcal{D}^\text{OM}_\text{out}$ by selecting the top $N$ samples with the largest $\hat{G}(\*x)$ from  $\mathcal{D}_{\text{pool}}$\\
 	\tcp{update weights in the backbone and multi-class classification branch}
	Train for one epoch with energy-regularized cross-entropy loss $ L_\text{cls} + \beta L_\text{reg}$ defined in Equation~(\ref{eq:fine-tune-obj})\\
	\tcp{use queue to save computational cost}
	Update the feature queue $Q = \{\phi(\*x_i)\}_{i=1}^M$ based on  $\mathcal{D}^\text{OM}_\text{out}$ and $\mathcal{D}_\text{in}^{\text{train}}$\\
	 	\tcp{update feature representation matrix based on the queue}
	Update  $\Phi = [\phi(\*x_1)| \phi(\*x_2)| \ldots | \phi(\*x_M)]$ by concatenation \\
 	\tcp{the posterior is updated to facilitate exploration}
	Update the posterior covariance matrix $\Sigma_p  = \sigma^{-2}\Phi\Phi^\top + \Sigma^{-1}$ \\
 	\tcp{sample from the posterior distribution}
	Sample $\*w_{t+1}  \sim  \mathcal{N}\left( \sigma^{-2} \Sigma_p^{-1}\Phi  {\*y_{\text{tar}}}, \Sigma_p^{-1}\right)$
	}
\caption{Posterior Sampling-based Outlier Mining (POEM)}
\label{alg:BLR}
\end{algorithm*}}

\subsection{Approximate Posterior with Neural Networks}
\label{sec:posterior}
A key challenge in our outlier mining framework is \emph{how to maintain and update the posterior over models}? Because the exact posterior is often
intractable, we need ways to approximate them efficiently. Recent advances in approximate
Bayesian methods have made posterior approximation for flexible neural network
models practical. As shown in \citet{riquelme2018deep}, performing Bayesian linear regression on top of feature representations extracted via a neural network is tractable, easy-to-tune, and enjoys strong performance compared to a wide range of other Bayesian approximations.  This also circumvents the problem of linear algorithms due to their lack of representational power.
\paragraph{Bayesian linear regression based on deep features.} 
Give the above considerations, in this work, we choose to perform Bayesian linear regression (BLR) on top of the penultimate layer feature $\phi(\*x) \in \mathbb{R}^m$ of a deep neural network to model the boundary score:
$$\hat{f}_\text{outlier}(\*x;\*w_t) =\*w_t^\top \phi(\*x),$$
where $\*w_t \in \mathbb{R}^m$ is the weight parameter sampled from  a posterior distribution (\emph{c.f.} Equation~\ref{eq1}). Based on the sampled model $\*w_t$, we can perform outlier mining by selecting a set of outliers from the auxiliary dataset according to the estimated boundary score $\hat{G}(\*x)=-|\hat f_\text{outlier}(\*x; \*w_t)|$. We use the notation $\hat{f}_\text{outlier}(\*x;\*w_t)$ to distinguish model estimation from the groundtruth function ${f}_\text{outlier}(\*x;\*w^*)$ where $\*w^*$ is unknown. The target logits are set similar to Weber et~al.~\yrcite{weber2018optimizing}; see Section~\ref{dataset} for details. 

Besides simplicity and tractability, we will later show in Section~\ref{result} that the above model based on deep features demonstrates strong empirical performance across a wide range of tasks and enjoys good theoretical properties.

\paragraph{Tractable and efficient posterior update.} We now describe details on the distribution of $\*w$. One benefit of Bayesian linear regression is the existence of closed-form formulas
for posterior update~\cite{rasmussen2003gaussian}. We assume a Gaussian prior of $\*w_0\sim\mathcal{N}(0, \Sigma)$. The posterior distribution of $\*w_t$ is a multivariate Gaussian with the following closed form: 
\begin{equation}
\label{eq1}
    \*w_t \sim  \mathcal{N}\left( \sigma^{-2} \Sigma_{\text{p}}^{-1}\Phi  {\*y_{\text{tar}}}, \Sigma_{\text{p}}^{-1}\right),
\end{equation}
where $\Sigma_{\text{p}} := \sigma^{-2}\Phi\Phi^\top + \Sigma^{-1}$ is the posterior covariance matrix, $\Phi\in\mathbb{R}^{m\times M}$ is the concatenation of feature representations $\lbrace\phi(\mathbf{x}_i)\rbrace_{i=1}^{M}$, and $  {\*y_{\text{tar}}}\in\mathbb{R}^{M}$ is the concatenation of target logit values, and $\sigma^2$ the variance of \emph{i.i.d.} noises in target logit values. In practice, we use a fix-size queue to store $M$ most recent feature vectors $\Phi$ to save computation. $M$ can be much smaller than the auxiliary dataset size $M \ll |\mathcal{D}_\text{aux}|$. Similar techniques are used in~\citet{azizzadenesheli2018efficient}. We show in Section~\ref{tradeoff} that our method overall achieves strong performance with computational efficiency. 

\vspace{0.2cm}
\textbf{Interleaving posterior update with feature update.}
During training, we interleave posterior update with the feature representation update, which are mutually beneficial as the former helps select near-boundary outliers that facilitate learning a good representation, which in turn improves the decision boundary estimation. 
Specifically, at the start of each epoch, we first sample the model from the posterior, based on which we select a set of outliers from the auxiliary dataset according to the \emph{estimated} boundary score. Then we train the neural network for one epoch and update a fix-size queue with new features, based on which we update the posterior distribution of the model. We present the pseudo-code of POEM in Algorithm \ref{alg:BLR}.

\vspace{0.2cm}
\textbf{Remark: Justification for choosing Thompson sampling with BLR.} We choose Thompson sampling with BLR because it is simple and effective in practice, and also enjoys good theoretical guarantees. \textbf{(1)} Empirically, due to the stochastic nature of Thomspon sampling, there is no need for extra hyper-parameters except for the prior distribution. In other statistical optimization methods like upper confidence bound (UCB~\cite{auer2002using}), the upper bound is another critical hyper-parameter, and choosing an upper bound to cover the actual value with high probability can be hard in some cases \cite{russo2014learning}. Sub-optimal construction of such upper bound can lead to a lack of statistical efficiency \cite{osband17a}. \textbf{(2)} Theoretically, Thompson sampling with BLR matches the regret bound of linear UCB from a Bayesian view \cite{russo2014learning}, while it also enjoys computational tractability. Other methods like entropy-based methods, probability matching, and Bayesian neural networks do not enjoy such strong guarantees to our knowledge. In one sentence, Thompson sampling with BLR is a good trade-off between computational tractability and OOD detectability.

\subsection{Putting Together: Learning and Inference} 
\label{sec:train}
Lastly, we address the third challenge: \emph{how can we leverage the mined outliers for model regularization}? We present the full training and evaluation algorithm for multi-class classification and OOD detection.

\textbf{Architecture.} As shown in Figure~\ref{fig_arch}, our framework consists of two branches: a \emph{classification branch} (in green) with $K$ outputs of class labels, and an \emph{outlier mining} branch (in beige) where Bayesian neural linear regression is performed. Two branches share the same feature representation. The feature vector $\phi(\*x)\in \mathbb{R}^m$ goes through a linear transformation with weight matrix $\*W_h \in \mathbb{R}^{m\times K}$, followed by a softmax function. The final softmax prediction is a vector $F(\*x)  = \text{softmax}(\*W_h^\top\cdot \phi(\*x))\in \mathbb{R}^K$.

\textbf{Training procedure.} The overall training workflow consists of three steps: \textbf{(1)} Constructing an auxiliary outlier training set by selecting outliers with the highest \emph{estimated} boundary scores from a large candidate pool. \textbf{(2)} The classification branch together with the network backbone are trained using a mixture of ID and selected outlier data. \textbf{(3)} Based on the updated feature representation, we perform the posterior update of the weights in the outlier mining branch. The pseudo-code is provided in Algorithm~\ref{alg:BLR}. We proceed by describing the training objective for the classification branch.

\textbf{Training objective.} Our overall training objective is a combination of standard cross-entropy loss, together with a regularization term $L_\text{reg}$:
\begin{align}\label{eq:fine-tune-obj}
\min_\theta & \quad \mathbb{E}_{(\*x,c)\sim \mathcal{D}_{\text{in}}^{\text{train}}}[-\log {F_c}(\*x;\theta)] + \beta \cdot L_\text{reg},
\end{align}
where $\mathcal{D}_{\text{in}}^{\text{train}}$ is the in-distribution training data and $\theta$ is the parameterization of the neural network. We leverage the energy-regularized learning~\cite{liu2020energybased}, which performs the classification task while regularizing the model to produce \emph{lower} energy for ID data and \emph{higher} energy for the auxiliary outliers. This helps create a strong energy gap that facilitates test-time OOD detection. The regularization is defined by:
\begin{align}
  L_\text{reg} & = \mathbb{E}_{\*x_\text{in}\sim \mathcal{D}_{\text{in}}^{\text{train}}} (\max(0, E(\*x_\text{in};\theta)-m_{\text{in}}))^2 \\
  & + \mathbb{E}_{\*x_\text{out}\sim \mathcal{D}_{\text{out}}^{\text{OM}}} (\max(0, m_{\text{out}}-E(\*x_\text{out};\theta)))^2,
\end{align}
where $\mathcal{D}_{\text{out}}^{\text{OM}}$ is the auxiliary outliers selected by our outlier mining procedure (Section~\ref{sec:ts}). $m_\text{in}$ and $m_\text{out}$ are margin hyperparameters, and the energy ${E}(\mathbf{x};\theta) :=-\log \sum_{i=1}^K e^{f_i(\mathbf{x};\theta)}$ as defined in \citet{liu2020energybased}. While \citet{liu2020energybased}  randomly sampled outliers from the pool, we perform outlier mining based on posterior sampling. We will show in Section~\ref{experiment} that OOD detection performance can be improved significantly with our novel outlier selection.

\paragraph{OOD inference.} At test time, OOD detection is based on the energy of the input: $D_\lambda(\mathbf{x}) = \mathbbm{1}\{-{E}(\mathbf{x}) \ge \gamma\}$,
where a threshold mechanism is exercised to distinguish between ID and OOD. Note that we negate the sign of the energy $E(\*x)$ to align with the convention that 
samples with higher scores are classified as ID and vice versa. The threshold $\gamma$ is typically chosen so that a high fraction of ID data (\emph{e.g.,} 95\%) is correctly classified.

\begin{table*}[t]
\centering
\caption[]{\small \textbf{Main Results}. Comparison with competitive OOD detection methods trained with the same DenseNet backbone. All values are percentages and are averaged over six OOD test datasets described in Section~\ref{dataset}. Bold numbers indicate superior results. We report the performance of POEM based on 5 independent training runs. Standard deviations for competitive baselines can be seen in Figure~\ref{diff_epoch}.}

\vskip 0.08in
\scalebox{0.82}{
\begin{tabular}{clllllcc}
\toprule
$\mathcal{D}_\text{in}$                      & \textbf{Method}               & \textbf{FPR95}$\downarrow$   & \textbf{AUROC}$\uparrow$  & \textbf{AUPR}$\uparrow$ & ID-ACC & w./w.o. $\mathcal{D}_\text{aux}$ &Sampling Method\\ \midrule
\multirow{10}{*}{\textbf{CIFAR-10}}   
                        & MSP \citep{MSP}                   & 58.98    & 90.63  & 93.18  & \textbf{94.39} & \xmark &NA\\ 
                        & ODIN   \citep{ODIN}                & 26.55   & 94.25 &  95.34  &94.39 & \xmark&NA\\ 
                        & Mahalanobis \citep{Maha}                 & 29.47  & 89.96 & 89.70 &94.39  & \xmark&NA\\ 
                        & Energy    \citep{liu2020energybased}               & 28.53    &94.39  & 95.56&94.39  & \xmark&NA\\ 
                         &  SSD+~\citep{2021ssd} & 7.22 & 98.48 & 98.59 & NA  & \xmark&NA\\
                        & OE    \citep{OE}               & 9.66   & 98.34 & 98.55 & 94.12 & \cmark& random\\ 
                          & SOFL   \citep{Self}              & 5.41   & 98.98 &  99.10 &93.68 & \cmark & random\\ 
                          & CCU   \citep{ccu}               & 8.78    & 98.41 &98.69  &93.97 &\cmark& random\\ 
                          & NTOM  \citep{chen2020informative}     & 4.38  & 99.08 & 99.24 &94.11 & \cmark& greedy\\ 
                        & Energy (w. $\mathcal{D}_\text{aux}$) \citep{liu2020energybased}             & 4.62    & 98.93 & 99.12 &92.92 & \cmark& random\\
                          &   \textbf{POEM} (ours) & $\textbf{2.54}^{\pm0.56}$   & $\textbf{99.40}^{\pm0.05}$   & $\textbf{99.50}^{\pm0.07}$  &$93.49^{\pm0.27}$ & \cmark& Thompson\\ 
                          \midrule
\multirow{10}{*}{\textbf{CIFAR-100}} 
                        & MSP  \citep{MSP}                  & 80.30   & 73.13 & 76.97 &\textbf{74.05}& \xmark&NA\\ 
                        & ODIN     \citep{ODIN}              & 56.31    & 84.89 & 85.88&74.05& \xmark&NA \\ 
                        & Mahalanobis  \citep{Maha}       & 47.89   & 85.71 & 87.15 &74.05 & \xmark&NA\\ 
                        & Energy     \citep{liu2020energybased}    & 65.87     &81.50  &84.07 &74.05& \xmark &NA\\ 
                         &  SSD+~\citep{2021ssd} & 38.32 & 88.91 & 89.77& NA  & \xmark&NA\\
                        & OE  \citep{OE}                 & 19.54   & 94.93 & 95.26  &74.25 & \cmark  &random\\ 
                          & SOFL   \citep{Self}               & 19.32  & 96.32 & 96.99 &73.93& \cmark& random \\ 
                          & CCU  \citep{ccu}                & 19.27  & 95.02  & 95.41 &74.49& \cmark& random\\ 
                          &  NTOM  \citep{chen2020informative}     & 19.96   & 96.29  & 97.06 &73.86& \cmark &greedy\\ 
                        & Energy (w. $\mathcal{D}_\text{aux}$) \citep{liu2020energybased}             & 19.25   & 96.68  &97.44  &72.39&\cmark&random \\
                          &  \textbf{POEM} (ours) & $\textbf{15.14}^{\pm1.16}$   & $\textbf{97.79}^{\pm0.17}$  &  $\textbf{98.31}^{\pm0.12}$  &$73.41^{\pm0.21}$& \cmark& Thompson\\
\bottomrule
\end{tabular}
}
\label{main_result}
\end{table*}

\section{Experiments}
\label{experiment}
In this section, we present extensive experiments to validate the superiority of POEM.  We also provide comparisons with three broad categories of OOD detection methods based on: (1) pre-trained models, (2) models trained with randomly sampled outliers, and (3) models trained with greedily sampled outliers. {Code is publicly available at: \url{https://github.com/deeplearning-wisc/poem}.}

\subsection{Experimental Setup}
\label{dataset}

\textbf{Datasets.} 
Following the common benchmarks, we use \texttt{CIFAR-10} and \texttt{CIFAR-100} \cite{krizhevsky2009learning} as in-distribution datasets. A downsampled version of ImageNet (\texttt{ImageNet-RC}) \cite{chrabaszcz2017downsampled} is used as the auxiliary outlier dataset. For OOD test sets, we use a suite of diverse image datasets including \texttt{SVHN} \cite{netzer2011reading}, \texttt{Textures} \cite{texture}, \texttt{Places365} \cite{zhou2017places}, \texttt{LSUN-crop}, \texttt{LSUN-resize}~\cite{yu2015lsun}, and \texttt{iSUN}~\cite{isun}.

\textbf{Training details.} The pool of outliers consists of randomly selected 400,000 samples from ImageNet-RC, and only 50,000 samples (same size as the ID training set) are selected for training based on the boundary score. For efficiency, we follow common practice and maintain a queue with data in the previous 4 epochs for the posterior update \cite{weber2018optimizing}. In other words, the size of our queue (which is also the action space we select outliers from) is $M=50,000 \times 4$. We use DenseNet-101 as the backbone for all methods and train the model using stochastic gradient descent with Nesterov momentum \cite{duchi2011adaptive}. We set the momentum to be 0.9 and the weight decay coefficient to be $10^{-4}$. The batch size is $64$ for {both} in-distribution and outlier training data. Models are trained for 100 epochs. For a fair comparison, the above setting is the \emph{same} for all methods trained with outliers. SSD+~\cite{2021ssd} employed the supervised contrastive loss~\cite{2020supcon}, which requires a larger batch size (1024) and longer training time (400 epochs) than the softmax cross-entropy loss.
For margin hyperparameters, we use the default as in \citet{liu2020energybased}: $m_{\text{in}}= -7$, $m_{\text{out}} = -25$ and $\beta = 0.1$.

In Bayesian regression, we follow the practice in~\citet{weber2018optimizing}, and regress on target logit values of 3 (for $\*x \in \mathcal{D}_\text{aux}$, corresponding to a target probability of 0.95) and -3 (for $\*x \in \mathcal{D}_\text{in}^\text{train}$ with a target probability of 0.05) with \emph{i.i.d.} noise $\upsilon \sim \mathcal{N}(0,\sigma^2)$. Note that we avoid regressing on 0-1 labels since it incurs infinite logits. As a result, the reward incurred is $|3+\upsilon|$ for $\*x \in \mathcal{D}_\text{aux}$ and $|-3+\upsilon|$ for $\*x \in\mathcal{D}_\text{in}^\text{train}$. In this setting, Thompson sampling can still find new OOD data that are closer to the boundary (see detailed explanation in Appendix~\ref{fixedlogit}).

\paragraph{Evaluation metrics. } Following common practice in the literature, we report: (1) the false positive rate (\text{FPR}95) of OOD samples when the true positive rate of in-distribution samples is at 95\%; (2) the area under the receiver operating characteristic curve (AUROC); (3) the area under the precision-recall curve (AUPR). We also report the ID classification accuracy (ID-ACC).

 \subsection{Results and Discussion} \label{result}
\textbf{{POEM} achieves SOTA performance.} Our method outperforms existing competitive methods, establishing \emph{state-of-the-art} performance on both CIFAR-10 and CIFAR-100. Table~\ref{main_result} summarizes detailed comparison with methods that \textbf{(1)} directly use a pre-trained network for OOD detection: \texttt{MSP} \cite{MSP}, \texttt{ODIN} \cite{ODIN}, \texttt{Mahalanobis} \cite{Maha}, \texttt{Energy}~\cite{liu2020energybased}; \textbf{(2)} use an auxiliary outlier dataset but randomly select outliers during training: \texttt{OE} \cite{OE}, \texttt{SOFL} \cite{Self}, \texttt{CCU} \cite{ccu}, energy-regularized learning (\emph{i.e.}, Energy (with outlier)); \textbf{(3)} involve outlier mining: \texttt{NTOM}~\cite{chen2020informative}. 
Compared to the best baseline energy-regularized learning, \texttt{POEM} reduces the FPR95 by 4.1\% on CIFAR-100, which is a relative \textbf{21.4}\% reduction in error. This highlights the benefits of posterior-sampling-based outlier mining, as opposed to randomly selecting outliers as in~\citet{liu2020energybased}.

\begin{figure*}[t]
  \centering
    \includegraphics[width=0.925\linewidth]{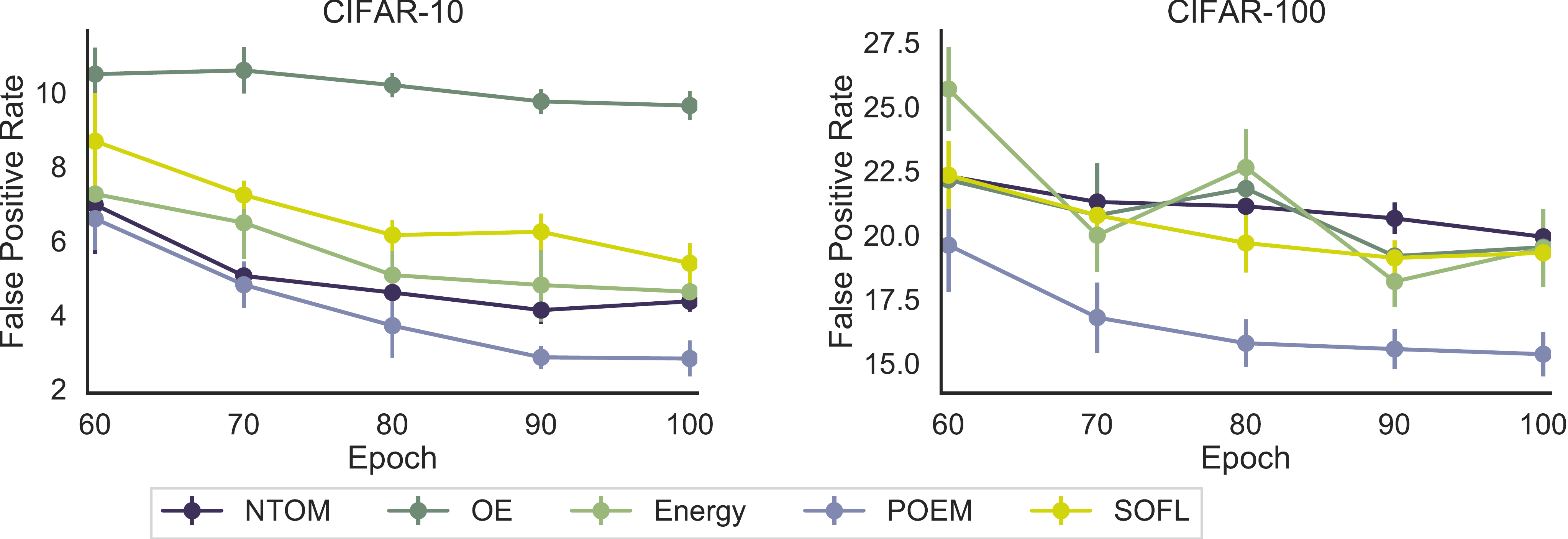}
      \vspace{-0.1cm}
\caption{\small Performance comparison of methods that utilize an auxiliary outlier dataset on CIFAR-10 (left) and CIFAR-100 (right) at different epochs during training. \textbf{Mean} and \textbf{standard deviations} are estimated across five independent training runs.} 
\vspace{-0.3cm}
\label{diff_epoch}
\end{figure*}

\paragraph{Thompson sampling v.s. greedy sampling.} \label{vs} \citet{chen2020informative} also perform outlier mining for OOD detection, which is the most relevant baseline\footnote{There are two variants in \citet{chen2020informative}: NTOM (for standard training) and ATOM (for adversarial training). NTOM has comparable performance as ATOM on natural datasets, and is a more fair comparison for our setting.}. \texttt{NTOM} can be viewed as a simple greedy sampling strategy, where outliers are selected based on the estimated confidence without considering the uncertainty of model parameters. Despite the simplicity, this method only exploits what is currently available to the model, while falling short of proper exploration. In contrast, our method achieves a better exploration-exploitation trade-off by maintaining a posterior distribution over models. Empirically, under the same configurations, our method outperforms \texttt{NTOM} by 4.82\% (FPR95) on CIFAR-100. Moreover, while the performance of NTOM can be sensitive to the confidence hyperparameter (see Table~\ref{ablation} in Appendix), our framework does not suffer from this issue. Since the estimated decision boundary is adjusted through the posterior update, \texttt{POEM} obviates the need for a dataset-dependent hyperparameter. 

\vspace{0.2cm}
\textbf{{POEM} utilizes outliers more effectively than existing approaches.} Figure~\ref{diff_epoch} contrasts the OOD detection performance of various methods at different training epochs. For a fair comparison, we mainly consider methods that use auxiliary outlier datasets. We report the average \text{FPR}95 (with standard deviation) at varying epochs. On both CIFAR-10 and CIFAR-100, \texttt{POEM} achieves lower FPR95 with fewer training epochs. Our experiments suggest that \texttt{POEM} utilizes outliers more effectively than existing approaches. This also highlights the importance of selectively choosing the outliers near the decision boundary between ID and OOD. 

\vspace{0.2cm}
\textbf{{POEM} improves OOD detection while maintaining comparable classification accuracy.}
We compare the multi-class classification accuracy in Table~\ref{main_result}. When trained on DenseNet with CIFAR-100 as ID, \texttt{POEM} achieves a test error of 26.59\%, compared to the NTOM fine-tuned model's 26.14\% and the pre-trained model's 25.95\%. Overall our training method leads to improved OOD detection performance, and at the same time maintains comparable classification accuracy on in-distribution data. 

{Due to space constraints, {additional ablation studies such as the impact of \emph{alternative auxiliary datasets}, and the \emph{pool size of auxiliary datasets} are included in Appendix \ref{other_ablation}.}}

\section{Further Discussion on Computation}
\label{tradeoff}
As with most Bayesian methods, there is no ``free lunch'': keeping track of the uncertainty of model parameters comes at a cost of additional computation. In this section, we discuss the computation and performance trade-off, and we demonstrate that our method \texttt{POEM} achieves strong performance with moderate computational cost. We also provide some practical techniques to further reduce computation. The estimated average runtime for each method (utilizing outliers) is summarized in Table~\ref{runtime}. Methods utilizing auxiliary data generally take longer to train, as a trade-off for superior detection performance. Among those, \texttt{SOFL}~\cite{Self} is the most computationally expensive baseline (14 hours). The training time of \texttt{POEM} is shorter than \texttt{SOFL}. Moreover, the {performance gain is substantial, establishing state-of-the-art performance}. For example, the average FPR95 is reduced from 19.25\% to 15.14\% on CIFAR-100. 
To put the numbers in context, compared with methods using outlier mining such as \texttt{NTOM} \cite{chen2020informative} (19.96\%), \texttt{POEM} still yields superior performance under comparable computations. 

We can improve the training efficiency of POEM using early stopping, \emph{i.e.}, {no outlier mining and posterior update} after certain epochs but regular training continues. The insight is taken from Figure~\ref{diff_epoch}, where \texttt{POEM} establishes SOTA performance around epoch 80. This further saves the average training time by {2.4 hours} with a marginal performance decrease (with AUROC 97.33\% on CIFAR-100).

\begin{table}[t]
\centering
\caption[]{\small The estimated average runtime for each method. Models are trained with DenseNet as the architecture. ImageNet-RC \citep{chrabaszcz2017downsampled} is used for methods that use an auxiliary outlier dataset. h denotes hour.  }
\vskip 0.08in
        \label{runtime}
\scalebox{0.9}{
                \begin{tabular}{lc}
                        \toprule
      \textbf{Method}  &\textbf{Training Time}   \\
 \hline
                              Energy (w. $\mathcal{D}_\text{aux}$)~\citep{liu2020energybased} & 5.0 h \\
                         CCU~\citep{ccu} & 6.7 h\\
                         NTOM~\citep{chen2020informative} & 7.4 h \\
                  SOFL~\citep{Self}  &  14.0 h\\
                         {POEM} ~&  10.3 h\\
                                                  {POEM} (early stopping)~&  7.9 h\\
 \bottomrule
                \end{tabular}
                }
        \vspace{-0.3cm}
\end{table}

\paragraph{Training with more randomly sampled outliers does not outperform POEM.} We verify that random sampling with more auxiliary samples does not yield more competitive performance compared to POEM. As shown in Table~\ref{tab:eff}, training \cite{liu2020energybased} with 3 times more auxiliary samples only marginally improved the performance (FPR95 from 19.25\% to 19.19\%, AUROC from 96.68\% to 97.18\%) on CIFAR-100. Here we refer to training with a larger outlier buffer with randomly sampled outliers. The overall size of the auxiliary dataset ImageNet-RC is kept the same. However, simply training with more randomly sampled outliers incurred a significant computational burden, where the training time increases from 5h to {8.9h}. The above observations further highlight the benefit of outlier mining with \texttt{POEM}.

\begin{table}[!h]
    \centering
    \footnotesize
       \caption[]{\small Training with more \emph{random} outliers does not improve the performance of Energy (w.  $\mathcal{D}_\text{aux}$)~\cite{liu2020energybased}. The results reported are the average of 5 independent runs.}
       \vskip 0.08in
\scalebox{0.95}{
\begin{tabular}{l|ccc}
\toprule
\textbf{Method} (CIFAR-100 as $\mathcal{D}_
\text{in}$) & \textbf{FPR95} \textbf{$\downarrow$} & \textbf{AUROC} \textbf{$\uparrow$} &  \textbf{Time}\textbf{$\downarrow$}\\  \midrule
  1x outliers (rand. sampling) & 19.25 & 96.68 & 5.0h\\
 3x outliers (rand. sampling) &19.19  & 97.18 & 8.9h  \\
 \bottomrule
\end{tabular}
}

    \label{tab:eff}
\end{table}

\section{Theoretical Insights: Sample Complexity with High Boundary Scores}
\label{theo}

\label{gmm}
In this section, we 
provide insights on how selecting data points with high boundary scores benefits sample efficiency under a simple Gaussian mixture model in binary classification. Due to space constraints, a discussion on more general models for OOD detection can be found in Appendix~\ref{extension}.

\textbf{Setup.} As our method works on the feature space, to simplify notations, we use $\*x$ to denote the extracted feature and the following distributions are defined on the feature space. We assume $\mathcal{P}_{\text{in}}=\mathcal{N}(\bm{\mu}, \sigma^2\bm{I})$, and $\mathcal{P}_{\text{aux}}= \mathcal{N}(-\bm{\mu},\sigma^2\bm{I})$, with $\bm{\mu}\in \mathbb{R}^d$.
The hypothesis class $\mathcal{H}=\{\text{sign}(\bm{\theta}^\top\*x),\bm{\theta}\in \mathbb{R}^d\}$, where a classifier outputs 1 if it predicts $\*x \sim \mathcal{P}_{\text{in}}$ and $-1$ if it predicts $\*x \sim \mathcal{P}_{\text{aux}}$. The overall feature distribution is a Gaussian mixture model with equal class priors: ${1\over 2}\mathcal{N}(\bm{\mu}, \sigma^2\bm{I})+{1\over 2}\mathcal{N}(-\bm{\mu},\sigma^2\bm{I}))$. 

\paragraph{Outliers with high boundary scores benefit sample complexity.}

We can write the false negative rate (\text{FNR}) and false positive rate (\text{FPR}) of our data model:
$\text{FNR}(\bm{\theta})=
erf(\frac{\bm{\mu}^\top\bm{\theta}}{\sigma||\bm{\theta}||}) = 
\text{FPR}(\bm{\theta}) $,
where $erf(x) = \frac{1}{\sqrt{2\pi}}\int_{x}^{\infty}e^{-t^2/2}dt$. Consider the classifier given by $\hat{\bm{\theta}}_{n,n'} = \frac{1}{n'+n} (\Sigma_{i=1}^{n'}\*x'_i-\Sigma_{i=1}^{n} \*x_i)$ where  each $\*x_i $ is drawn \emph{i.i.d.} from $ \mathcal{P}_{\text{aux}}$,  each $\*x'_i$ is drawn \emph{i.i.d.} from $ \mathcal{P}_{\text{in}}$. We have the following result:
\begin{theorem}
\label{thm} Assume the signal/noise ratio is large: $\frac{||\bm{\mu}||}{\sigma} = r_0 \gg 1$, the dimensionality/sample ratio is $ r_1 = \frac{d}{n}$, and we have access to data points $\*x \sim \mathcal{P}_{\text{aux}}$ that satisfy the following constraint of high boundary scores (on average):
\begin{equation*}
\frac{-\Sigma_{i=1}^n G(\mathbf{x}_i)}{n}\leq \epsilon.
\end{equation*}
There exists a constant $c$ that
\begin{equation}
\label{lb}
\frac{\bm{\mu}^\top\hat{\bm{\theta}}_{n,n'}}{\sigma||\hat{\bm{\theta}}_{n,n'}||}\geq \frac{||\bm{\mu}||^2-\sigma^{1/2}||\bm{\mu}||^{3/2}-\frac{\sigma^2\epsilon}{2}}{2\sqrt{\frac{\sigma^2}{n}(d+\frac{1}{\sigma})+||\bm{\mu}||^2}}
\end{equation}
with probability at least $1-(1+c)e^{-r_1n/8\sigma^2}-2e^{-{1\over 2} n||\bm{\mu}||/\sigma}$.
\end{theorem}
\vspace{-0.1cm}
\textbf{Remark.} 
The above result suggests that when $\epsilon$ decreases, the lower bound of $\frac{\bm{\mu}^\top\hat{\bm{\theta}}_{n,n'}}{\sigma||\hat{\bm{\theta}}_{n,n'}||}$ will increase. As $\text{FNR}(\hat{\bm{\theta}}_{n,n'})=\text{FPR}(\hat{\bm{\theta}}_{n,n'})=
erf(\frac{\bm{\mu}^\top\bm{\hat{\bm{\theta}}_{n,n'}}}{\sigma||\bm{\hat{\bm{\theta}}_{n,n'}}||})$, then the upper bound of $\text{FNR}(\hat{\bm{\theta}}_{n,n'})$ and $\text{FPR}(\hat{\bm{\theta}}_{n,n'})$ 
will decrease, which shows the benefit of outlier mining with high boundary scores (see Appendix \ref{theorem1} for details).

\section{Related Works}
\label{related}

\textbf{Thompson sampling.} Thompson Sampling~\cite{thompson1933likelihood}, also known as posterior sampling, is an elegant framework that balances exploration and exploitation for bandit problems. It is easy to implement and can be combined with neural networks~\cite{riquelme2018deep,zhang2021neural}. Recently, Thompson sampling enjoys popularity in a wide range of applications such as recommendation systems~\cite{kawale2015efficient}, marketing~\cite{schwartz2017customer}, and web optimization~\cite{hill2017efficient}. While the framework is widely known for bandits and reinforcement learning problems, we are the first to establish a formal connection between OOD detection and Thompson sampling.

\textbf{Gaussian process with Thompson sampling.} Optimizing an unknown reward function has been well studied. We follow a line of works that use Gaussian processes to model continuous functions. 
\citet{russo2014learning} explore Thompson sampling under the Gaussian process assumptions and develop a Bayesian regret bound. \citet{riquelme2018deep} investigate a range of bandit problems and show that linear Bayesian models on top of a two-layer neural network representation serve as a satisfying extension of Gaussian processes in higher dimensional spaces. \citet{azizzadenesheli2018efficient} and \citet{fan2021model} demonstrate the success of such methods in reinforcement learning.

\textbf{Out-of-distribution detection without auxiliary data.} In~\citet{bendale2016towards}, the OpenMax score
is first developed for detecting samples from outside training categories, using the extreme value theory (EVT). Subsequent work~\cite{MSP} proposed a baseline using maximum softmax probability, which is not suitable for OOD detection as theoretically shown in~\citet{morteza2022provable}.  Advanced techniques are proposed to improve the detection performance exploiting the logit space, including ODIN~\cite{ODIN}, energy score~\cite{liu2020energybased,lin2021mood, wang2021canmulti}, ReAct~\cite{sun2021react}, and logit normalization~\cite{wei2022mitigating}. Compared to max logit~\cite{vaze2021open}, the energy score enjoys theoretical interpretation from a log-likelihood perspective. OOD detection based on the feature space such as Mahalanobis distance-based score~\cite{Maha} and non-parametric KNN-based score~\cite{sun2022knn} also demonstrates promises, especially on OOD with spurious correlation~\cite{ming2022impact}. \citet{zhang2020hybrid} find that jointly learning a classifier with a flow-based density estimator is effective for OOD detection. Without pre-training on a large dataset~\cite{fort2021exploring}, the performance of post hoc OOD detection methods is generally inferior to those that use auxiliary datasets for model regularization.

\textbf{Out-of-distribution detection with auxiliary data.}
 Another line of work incorporates an auxiliary outlier dataset during training, which may consist of natural \cite{OE, Self,liu2020energybased, chen2020informative, katzsamuels2022training} or synthesized OOD training samples~\cite{lee2018gan,du2022vos, du2022unknown}.
 Recently, \citet{chen2020informative} propose a greedy confidence-based outlier mining method. Despite the simplicity, it falls short of exploration. 
 Unlike existing methods using random or greedy sampling, our posterior sampling-based framework allows balancing exploitation and exploration, which enjoys both empirical benefits and theoretical properties. 
 
\textbf{Out-of-distribution detection in natural language processing.} 
Distribution shifts in NLP can occur due to changes in topics and domains, unexpected user utterances, etc. Compared to early language models such as ConvNets and LSTMs, pre-trained Transformers~\cite{vaswani2017attention} are shown robust to distribution shifts and more effective at identifying OOD instances~\cite{hendrycks-etal-2020-pretrained,Podolskiy21,xu-etal-2021-unsupervised}. Various algorithmic solutions are proposed to handle OOD detection, including model ensembling~\cite{lietal2021kfolden}, data augmentation~\cite{chen2021gold, zhan2021out}, and contrastive learning~\cite{zhou2021contrastive,zeng2021modeling, jin2022towards}. Since our method does not depend on any assumption about the task domain, it also has the potential to be applied to NLP tasks.

\section{Conclusion and Outlook}
\label{conclusion}

In this paper, we propose a novel posterior sampling-based learning framework (POEM) that facilitates learning a more compact decision boundary between ID and OOD for improved detection. A key to our framework is finding the near-boundary outlier training examples for model regularization. We conduct extensive experiments and show that POEM establishes the state-of-the-art among competitive OOD detection methods.  
We also provide theoretical insights on why selecting outliers with high boundary scores improves OOD detection. We hope our research can raise more attention to a broader view of using posterior sampling approaches for OOD detection. 
\section*{Acknowledgements} 
Research was supported by funding from the Wisconsin Alumni Research Foundation (WARF).
\bibliography{example_paper}
\bibliographystyle{icml2022}

\newpage
\appendix
\onecolumn


\section{Discussion on the Choice of Auxiliary Dataset and Pool Size}
\label{other_ablation}
\paragraph{Discussion on the use of auxiliary outlier dataset.} The assumption of having access to outliers has been adopted in a large body of OOD detection work (\emph{c.f.} baselines considered). The key premise of outlier exposure approaches is the availability of a large and diverse auxiliary dataset. Inheriting this setting, the precise challenge lies in how to use them in a more effective way that facilitates OOD detection. Thus, our goal is to propose a more \textbf{data-efficient} way to identify the most informative outliers which outperforms greedy approaches (NTOM) and random sampling (OE, CCU, SOFL). As shown in Figure~\ref{diff_epoch}, the performance gap between POEM and all baselines with outlier exposure plateaus around epoch 85. Training even longer might lead to overfitting with diminishing gains w.r.t. OOD detection performance. This further signifies the importance of effective outlier mining of POEM, even when computation is unconstrained. The sample efficiency is also theoretically justified (\emph{c.f.} Theorem~\ref{thm}). When the auxiliary dataset is limited or very different from ID data, outlier exposure approaches \emph{in general} may not be the most suitable; there exist other approaches to consider such as synthesizing outliers~\cite{du2022vos}.

\paragraph{The choice of auxiliary outlier datasets.} POEM is consistently competitive when using a different auxiliary outlier dataset. Following common practice, with TinyImages~\cite{tiny} as the auxiliary dataset, the average FPR95 of POEM is \textbf{1.56}\% on CIFAR-10, as shown in Table~\ref{tab:alter}.

\begin{table}[!h]
    \centering
    \footnotesize
     \caption[]{The effect of training with TinyImages as the auxiliary outlier dataset.  All methods are trained with the same DenseNet backbone. All values are percentages and are averaged over six OOD test datasets described in Section~\ref{dataset}. Bold numbers indicate superior results. Each number is based on the average of 5 independent training runs.}
   \vskip 0.08in
\scalebox{1}{
\begin{tabular}{l|c}
\toprule
Method (CIFAR-10 as $\mathcal{D}_\text{in}$) & {FPR95} \textbf{$\downarrow$} \\  \midrule
 POEM  & \textbf{1.56} \\
   NTOM~\cite{chen2020informative}  &3.27 \\
   OE~\cite{OE} & 4.33 \\
  Energy (w. $\mathcal{D}_\text{aux}$)~\cite{liu2020energybased} & 3.38  \\
 \bottomrule
\end{tabular}
}
  
    \vspace{-0.3cm}
    \label{tab:alter}
\end{table}

\paragraph{Effects of the pool size.} Our experiments suggest that POEM's performance is \textbf{insensitive} to the auxiliary pool size. The average AUROC slightly decreased from 99.40 to 99.29 on CIFAR-10 (FPR95 increased from 2.54 to 4.23), as a result of decreasing the pool size by half. POEM remains competitive compared to other baseline methods.

\section{Discussion on Modeling the Outlier Class as the K+1-th Class}
\paragraph{Ablation on the training loss.} As discussed in Section~\ref{vs} ({POEM} vs. {NTOM}), our method differs from NTOM~\cite{chen2020informative} in terms of both training objectives as well as outlier mining strategy. To isolate the effect of the training objective, we conduct an ablation by using the same training objective as in NTOM, however replacing the confidence-based outlier mining with POEM. In particular, $f_\text{outlier}$ is taken as the $K+1$-th component of the classification output $f_{K+1}(\*x)$. The performance comparison is shown in Table~\ref{ablation}. Under the alternative training scheme (dubbed $K+1$ scheme), POEM outperforms NTOM on both CIFAR-10 and CIFAR-100. Our ablation also indicates the superiority of using the energy-regularized training objective, compared to using the $K+1$ training scheme. In particular, on CIFAR-100, the average FPR95 decreases by 1.97\% with our method than using cross-entropy loss with $K+1$ classes. 

\begin{table}[H]
\centering
\caption{Comparison between NTOM and POEM under different schemes. POEM (K+1 scheme) refers to POEM with the same training and inference scheme as in NTOM. All values are percentages and are averaged over six natural OOD test datasets. Bold numbers indicate superior results.}
  \vskip 0.08in
\scalebox{0.8}{
\begin{tabular}{clccc}
\toprule
$\mathcal{D}_{in}$& \textbf{Method}& \textbf{FPR95}$\downarrow$&\textbf{AUROC}$\uparrow$&\textbf{AUPR}$\uparrow$  \\ 
\midrule                
\multirow{4}{*}{CIFAR-10} & NTOM (q=$0.125$)   & $7.21$  & $98.64$ &  $98.52$  \\
                        & NTOM (q=0)       & 4.38   & 99.08 & 99.24  \\
                           & POEM (K+1 scheme)   & $2.81$   & $99.28$ &  $99.41$  \\ 
                            & \textbf{POEM} (ours)  & $\textbf{2.54}$  & $\textbf{99.40}$ & $\textbf{99.50}$  \\ 
                           \midrule
\multirow{4}{*}{CIFAR-100}  & NTOM (q=$0.125$)   & $23.06$   & $95.16$ & $96.25$ \\ 
                           & NTOM (q=0)       & $19.96$   & $96.29$ & $97.06$  \\
                           & POEM (K+1 scheme)    & $17.11$   & $96.87$ & $97.73$ \\
                           & \textbf{POEM} (ours)  & $\textbf{15.14}$   & $\textbf{97.79}$ & $\textbf{98.31}$ \\ 
                           \bottomrule 
\end{tabular}
}
\label{ablation}
\end{table}

\section{The Choice of Logit Values in BLR}
\label{fixedlogit}
In Thompson sampling, the key is to find data with the highest estimated boundary score, not that from target values. There is no easy way to get the ground-truth ID/OOD probability, so we adopt a fixed target value from $p=0.95$ with noise for all ID/OOD data.
But it does not prevent finding new OOD data closer to the estimated boundary: at each iteration, we select data with the estimated highest boundary score $\hat{G}(x)$ via Thompson sampling from the auxiliary set, which means selected query samples are closer to the estimated boundary than any OOD data in the training pool, \emph{i.e.}, the newly selected OOD data is always ``in the middle'' of current ID and OOD data in the training pool. As we update the OOD training pool with the newly selected OOD data, the selected OOD becomes closer to ID, and the estimated boundary approaches the real one via Thompson sampling (see Figure~\ref{fig:teaser} (c) - (e) for visualization).

\section{Results on Individual Datasets}
We provide reference results of POEM on each OOD dataset in Table~\ref{tab:individual}, based on the publicly available checkpoints.

\begin{table*}[th]
\centering
\resizebox{\textwidth}{!}{
\begin{tabular}{lcccccccccccccc}
\toprule
\multirow{3}{*}{\textbf{ID Dataset}} & \multicolumn{12}{c}{\textbf{OOD Dataset}}                                                                                                                & \multicolumn{2}{c}{\multirow{2}{*}{\textbf{Average} }} \\
                        & \multicolumn{2}{c}{LSUN-crop} & \multicolumn{2}{c}{Places365} & \multicolumn{2}{c}{LSUN-resize} & \multicolumn{2}{c}{iSUN} & \multicolumn{2}{c}{Texture} & \multicolumn{2}{c}{SVHN} & 
                        \multicolumn{2}{c}{}                     \\
                        \cmidrule(lr){2-3}\cmidrule(lr){4-5}\cmidrule(lr){6-7}\cmidrule(lr){8-9}\cmidrule(lr){10-11}\cmidrule(lr){12-13}\cmidrule(lr){14-15}
                        & \textbf{FPR$\downarrow$}         & \textbf{AUROC$\uparrow$}      & \textbf{FPR$\downarrow$}           & \textbf{AUROC$\uparrow$}         & \textbf{FPR$\downarrow$}          & \textbf{AUROC$\uparrow$}        & \textbf{FPR$\downarrow$}         & \textbf{AUROC$\uparrow$}      & \textbf{FPR$\downarrow$}          & \textbf{AUROC$\uparrow$}        &
                        \textbf{FPR$\downarrow$}          & \textbf{AUROC$\uparrow$}        &\textbf{FPR$\downarrow$}               & \textbf{AUROC$\uparrow$}           \\
                        \midrule
CIFAR-10  & 13.36&97.52&1.47 &99.40&0.00 &100.00&0.00 &100.00&0.12 &99.90&0.37 &99.63&2.55 &99.41              \\
CIFAR-100 &49.85&92.87&5.92&98.29&0.00&100.00&0.00&100.00& 1.10&99.57&15.52&97.36&12.06&98.01\\
\bottomrule
\end{tabular}
}
\caption{\small OOD detection results of POEM on each OOD dataset (based on DenseNet-101). }
\label{tab:individual}
\end{table*}

\section{Theoretical Insights on Sample Complexity with High Boundary Scores}
\label{theorem1}
In this section, we provide details on how sampling with high boundary scores benefits the sample complexity. We begin with a brief review of some of the key definitions and notations.

\paragraph{Definitions and notations.} Due to the representational power of deep neural networks,  similar to prior works~\citep{Maha,2021ssd}, we assume the extracted feature approximately follow a Gaussian mixture model (GMM) with equal class priors: ${1\over 2}\mathcal{N}(\bm{\mu}, \sigma^2\bm{I})+{1\over 2}\mathcal{N}(-\bm{\mu},\sigma^2\bm{I}))$, where $\mathcal{P}_{\text{in}}=\mathcal{N}(\bm{\mu}, \sigma^2\bm{I})$, and $\mathcal{P}_{\text{aux}}= \mathcal{N}(-\bm{\mu},\sigma^2\bm{I})$. The hypothesis class $\mathcal{H}=\{\text{sign}(\bm{\theta}^\top\*x),\bm{\theta}\in \mathbb{R}^d\}$, where a classifier outputs 1 if it predicts $\*x \sim \mathcal{P}_{\text{in}}$ and $-1$ if it predicts $\*x \sim \mathcal{P}_{\text{aux}}$. The boundary score $G(\*x) = -|f_{\text{outlier}}(\*x)|$ (the ground truth weight $\*w^*$ is omitted for clarity). The probability of being an outlier is given by $p(\text{outlier}|\*x) = \text{Sigmoid}(f_\text{outlier}(\*x))$.

\begin{lemma} 
\label{lemma} Assume the selected data points $\*x \sim \mathcal{P}_{\text{aux}}$ satisfy the following constraint of high boundary scores (on average):
\begin{equation}
\label{newcons}
\frac{-\Sigma_{i=1}^n G(\mathbf{x}_i)}{n}\leq \epsilon
\end{equation}

Then we have
\begin{equation}
\label{res}
\Sigma_{i=1}^n|2\mathbf{x}_i^\top\bm{\mu}|\leq n\sigma^2\epsilon
\end{equation}
\end{lemma}
\paragraph{Proof of Lemma \ref{lemma}.}
We first obtain the expression for $G(\*x)$ under the Gaussian mixture model described above.

By Bayes' rule, $p({\text{outlier}}|\*x)$ can be expressed as:
\begin{equation}
\label{bayes}
    \begin{split}
    p({\text{outlier}}|\*x) &=\frac{p(\*x|{\text{outlier}})p({\text{outlier}})}{p(\mathbf{x})}\\
    &=\frac{\frac{1}{\sqrt{(2\pi\sigma^2)^d}}e^{-\frac{1}{2\sigma^2}d_{\text{outlier}}(\mathbf{x})}}{\frac{1}{\sqrt{(2\pi\sigma^2)^d}}e^{-\frac{1}{2\sigma^2}d_{\text{in}}(\mathbf{x})}+\frac{1}{\sqrt{(2\pi\sigma^2)^d}}e^{-\frac{1}{2\sigma^2}d_{\text{outlier}}(\mathbf{x})}}\\
    &=\frac{1}{1+e^{-\frac{1}{2\sigma^2}(d_\text{outlier}(\mathbf{x})-d_{\text{in}}(\mathbf{x}))}},
    \end{split}
\end{equation}
where $d_{\text{in}}(\mathbf{x}):=(\mathbf{x}-\bm{\mu})^\top(\mathbf{x}-\bm{\mu})$ and $d_{\text{outlier}}(\mathbf{x}) := (\mathbf{x}+\bm{\mu})^\top(\mathbf{x}+\bm{\mu})$. 

Note that $p(\text{outlier}|x) = \frac{1}{1+e^{-f_\text{outlier}(\mathbf{x})}}$. Thus, 
$f_{\text{outlier}}(\mathbf{x})=\frac{1}{2\sigma^2}(d_{\text{outlier}}(\mathbf{x})-d_\text{in}(\mathbf{x})).$ 

Then we have:
\begin{equation}
\label{closed}
    \begin{split}
    G(\mathbf{x}) &= -|f_{\text{outlier}}(\mathbf{x})| =-\frac{1}{2\sigma^2}|d_{\text{outlier}}(\mathbf{x})-d_\text{in}(\mathbf{x})| =-\frac{1}{2\sigma^2} |(\*{x}-\bm{\mu})^\top(\*{x}-\bm{\mu})- (\*{x}+\bm{\mu})^\top(\*{x}+\bm{\mu})|=-\frac{1}{2\sigma^2}|4\*{x}^\top\bm{\mu}|. \\
    \end{split}
\end{equation}

Therefore, the boundary score constraint $\frac{-\Sigma_{i=1}^n G(\mathbf{\mathbf{x_i}})}{n}\leq\epsilon$ is translated to:
\begin{equation}
\label{constraint}
\Sigma_{i=1}^n|2\mathbf{x}_i^\top\bm{\mu}|\leq n\sigma^2\epsilon
\end{equation}
As $\max_{i\in n}|\mathbf{x}_i^\top\bm{\mu}|\leq\Sigma_{i=1}^n|\mathbf{x}_i^\top\bm{\mu}|$, given a fixed $n$, the selected samples can be seen as generated from $\mathcal{P}_{\text{aux}}$ with the constraint that all samples lie within the two hyperplanes in (\ref{constraint}).

\paragraph{Sample Complexity Analysis.} Now we show the benefit of such constraint in controlling the sample complexity.
Assume the signal/noise ratio is large: $\frac{||\bm{\mu}||}{\sigma} = r_0 \gg 1$, the dimensionaility/sample size ratio is $r_1 = \frac{d}{n}$, and $\epsilon \leq 1$ is some constant.

Recall that we consider the classifier given by $\hat{\bm{\theta}}_{n,n'} = \frac{1}{n'+n} (\Sigma_{i=1}^{n'}\*x'_i-\Sigma_{i=1}^{n} \*x_i)$ where  each $\*x_i $ is drawn \emph{i.i.d.} from $ \mathcal{P}_{\text{aux}}$ and $\*x'_i$ is drawn \emph{i.i.d.} from $ \mathcal{P}_{\text{in}}$.
We can decompose $\hat{\bm{\theta}}_{n,n'} $ as:
\begin{equation}
    \hat{\bm{\theta}}_{n,n'} =\bm{\mu}+ \frac{n'}{n+n'}\bm{\theta}_1+\frac{n}{n+n'}\bm{\theta}_2,
\end{equation}
where $\bm{\theta}_1=\frac{1}{n'}(\Sigma_{i=1}^{n'} \*x'_i) -\bm{\mu} \sim \mathcal{N}(0,\frac{\sigma^2}{n'}I)$ and $\bm{\theta}_2=\frac{1}{n}(-\Sigma_{i=1}^{n} \mathbf{x}_i)-\bm{\mu}$.

For $\bm{\theta}_1$, we have that $||\bm{\theta}_1||^2 \sim \frac{\sigma^2}{n'}\chi^2_d$, and $\frac{\bm{\mu}^\top\bm{\theta}_1}{||\bm{\mu}||}\sim \mathcal{N}(0,\frac{\sigma^2}{n'})$. Then from standard concentration bounds: 
\begin{equation*}
    \begin{split}
        \mathbb{P}(||\bm{\theta}_1||^2\geq \frac{\sigma^2}{n'}(d+\frac{1}{\sigma}))&\leq e^{-d/8\sigma^2},\\
        \mathbb{P}(\frac{|\bm{\mu}^\top\bm{\theta}_1|}{||\bm{\mu}||}\geq (\sigma||\bm{\mu}||)^{1/2} )&\leq 2e^{-{1\over 2} n'||\bm{\mu}||/\sigma}.
    \end{split}
\end{equation*}
Since we have the equal prior probability for each class, we assume that $n=n'$. For $||\bm{\theta}_2||$,  since all $\mathbf{x}_i$ is drawn \emph{i.i.d.} from $\mathcal{P}_{\text{aux}}$ under the constraint (\ref{constraint}), so the distribution of $\bm{\theta}_2$ can be seen as a truncated distribution of $\bm{\theta}_1$. 
Thus, we have $\mathbb{P}(||\bm{\theta}_2||^2\geq \frac{\sigma^2}{n}(d+\frac{1}{\sigma}))\leq c\mathbb{P}(||\bm{\theta}_1||^2\geq \frac{\sigma^2}{n}(d+\frac{1}{\sigma}))\leq ce^{-d/8\sigma^2}$, where $c$ is some finite positive constant, $n'$ is replaced with $n$ in the above inequality.

The benefit of high boundary scores is brought by our constraint about $\bm{\theta}_2$: for $\bm{\mu}^\top\bm{\theta}_2$, unlike the analysis of $\bm{\mu}^\top\bm{\theta}_1$, we use results from Lemma~\ref{lemma} to obtain that $\frac{1}{n}\Sigma_{i=1}^n|\mathbf{x}_i^\top\bm{\mu}| \leq \frac{\sigma^2 \epsilon}{2}$ always holds. So $|\bm{\mu}^\top\bm{\theta}_2| \leq  ||\bm{\mu}||^2+\frac{\sigma^2 \epsilon}{2}$.

Now we can develop a lower bound for $\frac{\bm{\mu}^\top\hat{\bm{\theta}}_{n,n'}}{\sigma||\hat{\bm{\theta}}_{n,n'}||}$.
Let
\begin{equation}
||\bm{\theta}_1||^2\leq \frac{\sigma^2}{n}(d+\frac{1}{\sigma}) ,||\bm{\theta}_2||^2\leq \frac{\sigma^2}{n}(d+\frac{1}{\sigma}), \frac{|\bm{\mu}^\top\bm{\theta}_1|}{||\bm{\mu}||}\leq (\sigma||\bm{\mu}||)^{1/2}
\end{equation}hold simultaneously, we have $||\hat{\bm{\theta}}_{n,n'}||^2\leq \frac{\sigma^2}{n}(d+\frac{1}{\sigma})+||\bm{\mu}||^2$, and $|\bm{\mu}^\top\hat{\bm{\theta}}_{n,n'}| \geq \frac{1}{2}(||\bm{\mu}||^2-\sigma^{1/2}||\bm{\mu}||^{3/2}-\frac{\sigma^2\epsilon}{2})$.

Recall that $\frac{||\bm{\mu}||}{\sigma} = r \gg 1$, $\epsilon \leq 1$, then it satisfies that $||\bm{\mu}||^2-\sigma^{1/2}||\bm{\mu}||^{3/2}-\frac{\sigma^2\epsilon}{2} = \sigma^2(r^2-r^{3/2}-\frac{1}{2}) > 0$.

Thus via union bound, we have that
\begin{equation}
\label{lb}
\frac{\bm{\mu}^\top\hat{\bm{\theta}}_{n,n'}}{\sigma||\hat{\bm{\theta}}_{n,n'}||}\geq \frac{||\bm{\mu}||^2-\sigma^{1/2}||\bm{\mu}||^{3/2}-\frac{\sigma^2\epsilon}{2}}{2\sqrt{\frac{\sigma^2}{n}(d+\frac{1}{\sigma})+||\bm{\mu}||^2}}
\end{equation}

with probability at least $1-(1+c)e^{-d/8\sigma^2}-2e^{-{1\over 2} n||\bm{\mu}||/\sigma} = 1-(1+c)e^{-r_1n/8\sigma^2}-2e^{-{1\over 2} n||\bm{\mu}||/\sigma}$.

\paragraph{Interpretations.} The above results suggests that when $\epsilon$ decreases, the lower bound of $\frac{\bm{\mu}^\top\hat{\bm{\theta}}_{n,n'}}{\sigma||\hat{\bm{\theta}}_{n,n'}||}$ will increase. Recall that $\text{FNR}(\hat{\bm{\theta}}_{n,n'})=\text{FPR}(\hat{\bm{\theta}}_{n,n'})=
erf(\frac{\bm{\mu}^\top\bm{\hat{\bm{\theta}}_{n,n'}}}{\sigma||\bm{\hat{\bm{\theta}}_{n,n'}}||})$, so the upper bound of $\text{FNR}(\hat{\bm{\theta}}_{n,n'})$ and $\text{FPR}(\hat{\bm{\theta}}_{n,n'})$ 
will decrease accordingly, which shows the benefit of outlier mining with high boundary scores.

\section{Extension: Towards a More General Data Model for OOD Detection}
\label{extension}

In this section, we extend the data model in Appendix~\ref{theorem1} to a more general case where both the auxiliary and in-distribution data points are generated from mixtures of Gaussian distributions. Moreover, we consider a broader test OOD distribution $\mathcal{Q}_{\bm{v}}$ beyond $\mathcal{P}_{\text{aux}}$. We aim to provide more insights into the benefits of sampling with high boundary scores.

Specifically, the auxiliary data is generated with the following procedure: first draw a scalar $g$ from a uniform distribution with support $\{g: |g|\leq \frac{1}{4}||\bm{\mu}||_2\}$, then draw $\*x$ from $\mathcal{N}((-1+g)\bm{\mu},\sigma^2\bm{I})$. 
In addition, the in-distribution data points are generated as follows: first draw a scalar $s$ from a uniform distribution with support $\{s: |s|\leq \frac{1}{4}||\bm{\mu}||_2\}$, then draw $\*x$ from $\mathcal{N}((1+s)\bm{\mu},\sigma^2\bm{I})$. For the test OOD distribution, consider any $\mathcal{Q}_{\bm{v}}: \mathcal{N}(-\bm{\mu}+\bm{v},\sigma^2\bm{I})$, where $\bm{v} \in \mathbb{R}^d$ and $||\bm{v}||_2\leq \frac{1}{4}||\bm{\mu}||_2$.  

We can express the FPR as follows: 
\begin{equation}
\begin{split}
    \text{FPR}(\bm{\theta}) &= \mathbb{P}_{\*x \sim \mathcal{Q}_{\bm{v}}}(\bm{\theta}^\top \*x\geq 0)\\ &=\mathbb{P}(\mathcal{N}((-\bm{\mu}+\bm{v})^\top\bm{\theta},\sigma||\bm{\theta}||_2)\geq 0)\\&= erf(\frac{(\bm{\mu}-\bm{v})^\top\bm{\theta}}{\sigma||\bm{\theta}||_2})\\&\leq erf(\frac{\bm{\mu}^\top\bm{\theta}}{\sigma||\bm{\theta}||_2}-\frac{||\bm{\mu}||_2}{4\sigma}).
    \end{split}
\end{equation}
Consider any $\mathcal{P}_{\text{in}} = \mathcal{N}((1+s)\bm{\mu},\sigma^2\bm{I})$ where $|s|\leq \frac{1}{4}||\bm{\mu}||_2$ as the ID test set where we calculate the \text{FNR}. Similarly, we have $\text{FNR}(\bm{\theta})\leq erf(\frac{\bm{\mu}^\top\bm{\theta}}{\sigma||\bm{\theta}||_2}-\frac{||\bm{\mu}||_2}{4\sigma})$.

However, under such general data model, unlike in Appendix~\ref{theorem1}, there is no clean solution for $G(\*x)$ as equation (\ref{closed}): the probability density functions for $\mathcal{P}_{\text{in}}$ and $\mathcal{P}_{\text{aux}}$ require integrating Gaussian densities on w.r.t. the probability measure of $s$ and $g$, respectively. Thus when calculating the inverse of the Sigmoid function in $p(\text{outlier}|x) = \frac{1}{1+e^{-f_\text{outlier}(\mathbf{x})}}$ using the new probabilistic model, we cannot directly disentangle  $\*x^\top\mu$ from the expression of $f_{\text{outlier}}$ as equation (\ref{closed}). Nevertheless, we can expect that the decision boundary for $\mathcal{P}_{\text{in}}$ and $\mathcal{P}_{\text{aux}}$ is still a hyperplane under this setting. Recall that $p(\text{outlier}|\*x) = \frac{1}{1+e^{-f_\text{outlier}(\mathbf{x})}}$, which indicates that the property of data points with high boundary scores ( $p(\text{outlier}|\*x)\approx0.5$ ) can generally bound the distances between selected data and the decision hyperplane, so there exists two hyperplanes that can bound those auxiliary points we select with constraint (\ref{newcons}). As a result,  we will have a revised version of Lemma~\ref{lemma} with a different bound on the right hand side of equation (\ref{constraint}) which will still 
be positively correlated with $\epsilon$. 
Using the subgaussian property of $\bm{v}$ and
following similar steps in Appendix~\ref{theorem1}, we can expect similar sample complexity results with a different (but still negative) weight on $\epsilon$ compared to the results in Theorem~\ref{thm} which still shed light on the benefits of sampling with high boundary scores.

\section{Hardware and Software}
We run all the experiments on  NVIDIA GeForce RTX-2080Ti GPU. Our implementations are based on Ubuntu Linux 20.04 with Python 3.8.

\end{document}